\documentclass{article} % For LaTeX2e
\usepackage{iclr2027_conference,times}

\usepackage{amsmath,amsfonts,bm}

\def\eqref#1{equation~\ref{#1}}
\def\1{\bm{1}}

\DeclareMathAlphabet{\mathsfit}{\encodingdefault}{\sfdefault}{m}{sl}
\SetMathAlphabet{\mathsfit}{bold}{\encodingdefault}{\sfdefault}{bx}{n}

\usepackage{hyperref}
\usepackage{url}
\usepackage{booktabs}
\usepackage{tabularx}
\usepackage{array}
\usepackage{xcolor}
\usepackage{multirow}
\usepackage{colortbl}
\usepackage{graphicx}
\usepackage{pifont}
\usepackage{subcaption}
\usepackage{caption}
\usepackage{wrapfig}
\usepackage[section]{placeins}
\usepackage{needspace}
\usepackage{float}
\usepackage[most]{tcolorbox}
\usepackage{enumitem}
\usepackage{tikz}
\usepackage{booktabs}
\usepackage{wrapfig}
\usepackage[table]{xcolor}
\usepackage{siunitx}

\definecolor{revisionblue}{HTML}{00A8E8}
\definecolor{revisionblue}{HTML}{000000}
\newcommand{\sigone}{\textsuperscript{\textcolor{revisionblue}{\dag}}}
\newcommand{\sigtwo}{\textsuperscript{\textcolor{revisionblue}{\ddag}}}
\newcommand{\sigthree}{\textsuperscript{\textcolor{revisionblue}{\S}}}

\definecolor{revisionblue}{HTML}{2667ff}
\definecolor{revisionblue}{HTML}{000000}
\newcommand{\blackcirc}[1]{%
  \tikz[baseline=-0.6ex]{
    \node[
      circle,
      fill=black,
      text=white,
      minimum size=1em,
      inner sep=0pt,
      font=\scriptsize\bfseries
    ] {#1};
  }%
}
\definecolor{lightbluehl}{RGB}{225,240,255}
\definecolor{lightyellowhl}{RGB}{255, 249, 196}

\newtcolorbox{promptbox}[2][]{
  enhanced, breakable, sharp corners,
  colback=white, colframe=black, boxrule=0.9pt,
  fonttitle=\bfseries\large, coltitle=white,
  attach boxed title to top left={yshift=-2.6mm, xshift=5mm},
  boxed title style={colback=black, colframe=black, sharp corners,
    boxrule=0pt, left=4pt, right=4pt, top=2pt, bottom=2pt},
  top=5mm, bottom=3mm, left=4mm, right=4mm,
  title={#2}, #1}
  
\definecolor{lightbluehl}{RGB}{225,240,255}
\definecolor{lightyellowhl}{RGB}{255, 249, 196}

\newcommand{\cmark}{\textcolor{green!70!black}{\ding{51}}}
\newcommand{\xmark}{\textcolor{red}{\ding{55}}}
\newcommand{\empath}{\textsc{Empath}}

\definecolor{realcolor}{HTML}{5474AE}
\definecolor{semirealcolor}{HTML}{97B1D1}
\definecolor{syntheticcolor}{HTML}{AF4832}

\title{Right Destination, Wrong Journey: Synthetic Counseling Dialogues Miss Human Interactional Dynamics}
\author{%
\normalfont
\textbf{Ziwei Gong}$^{1}$\thanks{Equal contribution.} \quad
\textbf{Yuchen Huang}$^{2, 3}$\footnotemark[1]\quad
\textbf{Nicholas Deas}$^{1}$ \quad 
\textbf{Melanie Subbiah}$^{1,4}$ \quad 
\textbf{Julia Hirschberg}$^{1}$ \\[6pt]
$^{1}$Columbia University \quad
$^{2}$Barnard College \quad
$^{3}$UCLA \quad
$^{4}$Brown University \\[3pt]
\texttt{\{sara.ziweigong, ndeas, m.subbiah, julia\}@columbia.edu} \quad \\\texttt{yuchenhuang16@g.ucla.edu }
}
\definecolor{mypurple}{HTML}{8B639B}

\iclrfinalcopy
\begin{document}
\raggedbottom
\maketitle
\lhead{Preprint}
\raggedbottom

\begin{abstract}
LLM-generated counseling dialogues are increasingly used for training and evaluating mental-health systems, yet existing evaluations focus largely on response quality and final outcomes rather than how support unfolds over time. We conduct a trajectory-centered evaluation of process fidelity across \textcolor{revisionblue}{ten} human, reconstructed, and synthetic counseling datasets. 
% \ndnote{Maybe just not updated yet, but flagging this still sounds like the framework is a primary contribution. Is it supposed to be?} \msnote{agreed, maybe: "We conduct a trajectory-centered evaluation of process faithfulness across..."}
Our results show that synthetic dialogues often reach plausible endpoints but differ from human support in recovery dynamics and turn-by-turn responsiveness. While generation design can shift global properties such as recovery timing and trajectory structure, contingent patient--counselor responsiveness remains consistently lower in fidelity across models and generation paradigms. Our experiments on controlled synthetic datasets show that targeted generation interventions can narrow the gap with real data on several trajectory-level measures, but do not fully recover interactional fidelity, with turn-level responsiveness remaining limited. We release three controlled synthetic datasets comprising 2,050 dialogues, together with annotations and the evaluation framework. 
% \msnote{Could add a note on whether your synthetic datasets improve on process faithfulness (I think they do?)}

% LLM-generated counseling dialogues are increasingly used for counselor training, mental-health system development, and agent evaluation.Yet existing evaluations focus largely on response quality and final outcomes, rather than whether synthetic conversations reproduce realistic emotional trajectories.
% To bridge this gap, we introduce a trajectory-centered evaluation
% framework that uses turn-level emotion states to measure how distress evolves over a conversation and how those changes respond to counselor behavior.
% We apply the framework to nine datasets from different sources including real, reconstructed, and synthetic dialogues produced under different protocols.
% Our results reveal that synthetic dialogues often reach realistic endpoints, but none reproduce the emotional trajectories of real conversations. In further analysis, we find that generation protocols can control what happens and when, but all synthetic corpora fail to produce the contingent, turn-by-turn responsiveness that defines real support. We test prompt-tuning techniques specifically designed to address these gaps. We also release three controlled synthetic datasets including 2,050 dialogues to support evaluation of counseling generation by the interaction trajectory. %\sara{TODO: add one sentence about prompt-tuning experiments here.} 

\end{abstract}

\section{Introduction}

Text-based emotional-support services offer low-barrier, anonymous help to people in distress. Because authentic counseling transcripts are sensitive, privacy-restricted, and costly to curate, LLM-generated dialogue has become an increasingly attractive substitute. Synthetic conversations are now used as training data for support models~\citep{lee-etal-2024-cactus,qiu-lan-2025-psydial}, simulated patients for counselor training~\citep{louie-etal-2024-roleplay,10.1145/3772318.3791821}, and benchmarks for evaluating counseling agents~\citep{pan-etal-2026-psycheval}. In each setting, the usefulness of a conversation depends not only on whether it flows smoothly or whether the patient feels better at the end, but also on how well the interaction and emotional trajectory resemble those of real mental health therapy. Recent evidence suggests that fluent and clinically structured synthetic
dialogues can nevertheless exhibit unrealistic emotional dynamics \citep{wang-etal-2025-feel}.

This distinction matters because counseling is inherently sequential and interactive. In real sessions, distress may persist, recur, or gradually ease, and the same emotional endpoint for the patient can arise through very different processes. In synthetic dialogues, however, relief may emerge gradually in response to the counselor or appear abruptly midway through the conversation as a consequence of the generation protocol. This creates a problem for model training: models trained on the latter may learn when recovery tends to occur without learning which counselor responses help produce it, thereby missing the critical turning points at which particular responses or strategies contribute to improvement.
% This distinction matters because counseling is inherently sequential and interactive: distress may persist, recur, or gradually ease, and similar endpoints can arise through very different trajectories. By a plausible destination, we mean a reasonable terminal state, such as a patient ending the conversation less distressed or more positive.  But the same endpoint may arise through very different processes: relief may emerge gradually in response to the counselor, or appear abruptly because the generation protocol schedules recovery near the end. Models trained on the latter may learn when recovery occurs without learning which responses help produce it, while trainees may receive little exposure to stalls, reversals, or weak responses to support. \ndnote{I'm a bit unclear on what "stalls" and "reversals" refers to.}\ndnote{Makes more sense to me after finishing, but not clear in intro} \msnote{I think you could simplify the start of this paragraph: "This distinction matters because counseling is inherently sequential and interactive: distress may persist, recur, or gradually ease, and the same emotional endpoint for the patient can arise through very different processes. Relief may emerge gradually in response to the counselor, or appear abruptly because the generation protocol schedules recovery near the end. Models trained on the latter may learn when recovery occurs without learning which responses help produce it, while trainees may receive little exposure to cases where progress plateaus, setbacks occur, or support doesn't seem to help."}
Existing evaluations rarely distinguish between these cases. Synthetic counseling
datasets are commonly evaluated using response-quality measures, clinical instruments
such as CTRS or WAI, downstream model utility, and session- or dialogue-level outcomes
\citep{lee-etal-2024-cactus,qiu-lan-2026-interactive,
qiu-lan-2025-psydial,pan-etal-2026-psycheval}. 
These evaluations capture important aspects of counselor technique, therapeutic
quality, and overall progression, and synthetic or reconstructed counseling data can match or even exceed real-data
references on conventional evaluation metrics such as working alliance, helper
skills, and downstream counselor performance
\citep{pan-etal-2026-psycheval,qiu-lan-2025-psydial}, while still differing from human dialogue in its interactional dynamics. 
\textcolor{revisionblue}{
Recent advances in reinforcement learning for psychological
counseling have further demonstrated the potential of
optimizing psychological reasoning, therapeutic strategy
selection, and context-sensitive response generation
\citep{dai-etal-2026-psyche,yang-etal-2026-mithinker,wang-etal-2025-flexible}.
These developments motivate complementary process-level
evaluation: beyond the quality of individual responses,
it is important to understand how counselor behavior
relates to patient emotional change over the course
of an interaction.
}
We therefore ask whether synthetic counseling data is faithful \textit{both} in destination and in the emotional and interactional journey that produces it. 
% For example, the fully scripted CACTUS corpus rates above human peer support on working alliance, helper skills, and session ratings~\citep[Tables 4 and 9]{ecnu2026psycheval}, while SimPsyDial reports higher scores than real counselor--patient sessions on every WAI subscale~\citep{qiu2024interactiveagents}. 

% \ndnote{I don't know what an alliance-level measure is}
We study this problem as process fidelity: whether a counseling dialogue reproduces not only plausible utterances and outcomes, but also the evolving trajectory of distress and its relationship to counselor behavior. 
% \lareina{change faithfulness to interactional realism would be better}
% \ndnote{Minor, but feel like "faithfulness" may make people think like factuality grounded in some particular conversation; I've liked fidelity elsewhere or something else to say like how much it looks like authentic conversations.}
% \sara{changed all to "process fidelity"}
We organize our study around three questions:\\
% \begin{center}
\emph{\textbf{RQ1:} Do LLM-generated counseling dialogues reproduce the emotional trajectory of real support?
% \msnote{This question assumes plausible outcomes are achieved (which again I'm still not sure what a plausible outcome is or how it is measured) and if you are emphasizing trajectories in this paper, I would simplify to "Do LLM-generated counseling dialogues reproduce the emotional trajectory of real support?"}\\
% Do synthetic and reconstructed counseling dialogues reproduce the emotional and interactional trajectories of human support, or only plausible outcomes?
\textbf{RQ2:} Which trajectory deviations vary with generation protocol, and which persist across them?\\
\textbf{RQ3:} Can targeted generation controls reduce the deviations we identify?}
% \msnote{RQ2 and RQ3 feel like overlapping questions to me, maybe something like RQ2: "Which trajectory deviations can be controlled through generation-protocol choices?", and RQ3: "Does model choice affect trajectory fidelity" Not sure if this is doable with how results are organized}
% \end{center}

To answer these questions, we introduce a trajectory-centered evaluation framework that analyzes counseling dialogue from local interaction to global conversation structure. We measure turn-level emotion and distress dynamics, counselor strategies and their relationship to subsequent patient states, recovery timing and structure, and trajectory diversity. We apply the framework to \textcolor{revisionblue}{ten} English and Mandarin datasets spanning human, reconstructed/semi-real, and fully synthetic dialogue, including three controlled synthetic corpora spanning paradigms and model families. This setup lets us compare not only real and synthetic data, but also how different construction choices shape the trajectories that synthetic data produces.

% \msnote{From here on focuses on results and contributions which feels a little long. Try to compress and add more visual structure in the contributions. The findings are interesting in the next paragraph but hard to parse without introducing more of these terms so I think they could be simplified to the high-level takeaways. Additionally, I think you can combine the findings with the methodological contributions (e.g., for contribution iii instead of saying you identify which aspects are easy to control, share the finding of what is easy vs. hard to control)}
% Across these datasets, we find a consistent gap. Synthetic dialogues can often reproduce plausible outcomes, but they do not reproduce the process by which those outcomes arise in human support. 
% Outcome- and timing-level properties, including ending sentiment, recovery onset, trajectory shape, archetype coverage, and strategy mix, vary substantially across datasets and often track identifiable generation choices. By contrast, responsiveness-level properties are much more persistent \msnote{not immediately clear whether persistent is good or bad, maybe "demonstrate less fidelity" or something that makes it clear it is a negative}: across model families and generation paradigms, synthetic patients show compressed turn-level responses to counselor behavior, and apparent counselor–patient coupling can often be explained by shared trajectory schedules or joint authorship rather than one speaker adapting to the other. 

Across these datasets, we find that synthetic dialogues can often reach plausible endpoints, exhibiting declining patient-distress trajectories that resemble the overall progress of real conversations. Yet turn-by-turn responsiveness remains consistently lower in fidelity across models and generation paradigms. Apparent patient--counselor coupling often reflects shared schedules or joint authorship rather than genuine adaptation. We further use these findings to examine which aspects of synthetic trajectories are sensitive to generation design choices. We will release our evaluation framework, code, and synthetic counseling dialogues in an anonymous repository. \footnote{\url{https://anonymous.4open.science/r/synthetic-counseling-data-D356}}

Our contributions are threefold:

\emph{i.} We introduce a trajectory-centered evaluation framework for counseling dialogue (\S\ref{sec:methods}) that measures process fidelity across turn-level emotional change, recovery structure, patient--counselor responsiveness, and trajectory diversity.
% We introduce a trajectory-centered evaluation framework of counseling dialogue (\S\ref{sec:methods}). We operationalize process faithfulness across multiple levels, complementing evaluations based only on response quality or final outcomes.

\emph{ii.} We conduct a broad comparison across \textcolor{revisionblue}{ten} human, reconstructed, and synthetic datasets, together with a controlled generation of 2,050 synthetic conversations across multiple paradigms and model families (\S\ref{sec:generation_conditions}). We show that synthetic conversations can match endpoint statistics while differing substantially in how recovery unfolds. 
% \ndnote{Particularly here; I immediately think "trajectories faithful to what?" as if there was a summary faithful to a source document, but this is different since we are saying trajectories that resemble authentic conversations.}
% We conduct broad comparison and analysis across ten human, reconstructed, and synthetic datasets, together with a controlled generation testbed of 2,050 conversations across multiple generation paradigms and model families (\S\ref{sec:generation_conditions}). 
 % released with generation conditions and per-utterance emotion and strategy labels. 

\emph{iii.} We use controlled prompt variants and targeted interventions to identify which aspects of synthetic trajectories are sensitive to generation design and which remain difficult to reproduce. Prompt- and protocol-level controls can shift outcomes, recovery timing, and other global trajectory properties toward real data, but do not necessarily recover realistic contingent turn-by-turn interaction.
% \ndnote{I don't think worth changing anything, but might just be prepared for pushback on any claims about what prompting can't fix, for example; can always get a critique that we didn't try X prompt. Actually, I like the wording commented out below about "remain difficult to reproduce" which feels more defensible to me; up to y'all though}
% We identify which aspects of synthetic trajectories are controllable through generation design and which remain difficult to reproduce, and use these findings to provide empirical guidance for constructing more faithful synthetic counseling data.

\section{Related Work}
% \msnote{I think you should mention EMPATH somewhere here and that you build on it}
\label{sec:related}
\textbf{Synthetic Mental Health Data} have been proposed to address the scarcity, privacy constraints,
and access restrictions associated with real clinical and counseling data
\citep{wang-etal-2025-feel,mai-ehr,elemam-ehr}.
Within mental health NLP, construction strategies range from structured
LLM-based counseling dialogue synthesis, such as CATCH
\citep{chen-etal-2025-catch}, to hybrid resources such as MentalChat16K,
which combines synthetic counseling data with anonymized real interactions
\citep{xu2025mentalchat16k}. Related work in healthcare has also explored
fully synthetic health records for privacy-preserving data access
\citep{mai-ehr,elemam-ehr}.
These different construction strategies motivate our comparison across human,
reconstructed/semi-real, and fully synthetic counseling data.

% For example, the RealCBT dataset \citep{wang-etal-2025-feel} introduces synthetic therapy dialogues, but also found they lack the emotional variability and authentic reactivity of human interactions. To improve clinical fidelity, frameworks like CATCH \citep{chen-etal-2025-catch} utilize multi-agent chain-of-thought reasoning to mitigate "therapy drift". Hybrid datasets such as MentalChat16K \citep{MentalChat16K} offer another solution, blending AI-rephrased text with authentic clinician-youth interactions to preserve diagnostic complexity while protecting privacy. Synthetic Electronic Health Records \citep{mai-ehr,elemam-ehr} allows development and training of AI for analysis and other applications while maintaining privacy. \sara{add connection here on how we choose the datasets to compare in our paper.}
% \ndnote{(Nick to-do) find some citations that use the synthetic data for analysis of real data?}

\textbf{Patient Simulations} leverage language models as interactive virtual
patients for training mental-health practitioners and evaluating mental-health
systems. PATIENT-$\psi$ grounds simulated patients in CBT cognitive models
\citep{wang-etal-2024-patient}, while Eeyore improves simulation realism and
profile adherence through expert-guided instruction tuning and preference
optimization \citep{liu-etal-2025-eeyore}. Recent work further models
intervention-dependent patient state transitions
\citep{xu-etal-2026-puppet} or uses clinically grounded simulated users to
audit multi-turn chatbot behavior \citep{weilnhammer-etal-2026-simvail}. \textcolor{revisionblue}{More recently, multi-agent collaborative generation frameworks have become increasingly common. PsyChain \citep{feng-etal-2026-psychain} explored chain-of-agents framework to construct high-fidelity conseling dialogues through structured simulation. MultiAgentESC \citep{xu-etal-2025-multiagentesc} uses multiple LLM agents to collaboratively analyze emotional context, select support strategies, and evaluate candidate counselor responses without additional training, StoryMI \citep{meng-etal-2026-storymi} introduces a three-agent framework that grounds synthetic motivational interviewing dialogues in questionnaire-derived client narratives and dynamically controls therapeutic strategies through MI behavioral codes.} These efforts make the behavioral and interactional fidelity of simulated
patients an increasingly important assumption. 
% \msnote{There is other work on patient simulations for training/evaluating LLMs that I think would be worth citing to flesh out this section (e.g., https://www.nature.com/articles/s41591-026-04577-2.pdf}

% \textbf{Emotion Recognition in Conversation} is a foundational NLP task and key tool for tracking crisis intervention trajectories \citep{tripodi-etal-2025-assessing, Xu_2024}. In text-based therapeutic dialogues, model evaluations reveal clear tradeoffs: closed-source models like GPT-4 perform well in zero-shot diagnostic settings \citep{wu-etal-2025-multimodal, xie2025culturalpromptingimprovesempathy}. Open-weight LLaMA models, when lightly fine-tuned, achieve competitive performance in assessing emotional safety \citep{badawi-etal-2026-trust}. Domain-specific models consistently outperform general models on targeted clinical tasks  \citep{ji2022mentalbert}, while crisis-focused models leverage continuous valence scoring to track turn-level de-escalation in real-world crisis conversations \citep{tripodi-etal-2025-assessing}.

\textbf{Emotion Recognition} and sentiment modeling have been used to study
affective change and de-escalation in counseling and crisis-support dialogues
\citep{tripodi-etal-2025-assessing}. Prior work has also evaluated models for
emotion recognition, mental-health prediction, and therapeutic-response quality
\citep{wu-etal-2025-multimodal,ji2022mentalbert,badawi-etal-2026-trust,
xie2025culturalpromptingimprovesempathy}. More recently, RealCBT compares
emotional trajectories in authentic and synthetic therapy sessions
\citep{wang-etal-2025-feel}, while \empath{} examines emotion dynamics across
turns and conversation-level trajectories in crisis-support dialogues
\citep{gong2026empathtracingmultilevelemotion}. We extend this direction by
jointly evaluating emotional change, recovery structure, patient--counselor
responsiveness, and trajectory diversity.

% \lareina{found this paper which did very similar thing to our work with similar finding- but their scope is limited. I feel we need to cite them}

% \lareina{\citep{wang2025feel} find that LLM-generated CBT sessions differ from real
% therapy in emotional variability, reactivity, regulation, and trajectory
% alignment, particularly for simulated patients. We build on this observation
% by asking whether such deviations generalize across generation paradigms,
% which protocol choices produce them, and whether targeted generation controls
% can recover the interactional process of human support.}

%\sara{what are people thoughts about not having a seperate related work section? I feel like this related work section could be merged into intro+datasets} \msnote{I feel it's important to have.} \ndnote{I agree, I've had a lot of comments even when we put related work at the end of the paper so I'd just leave it. I feel like this is a good length/wouldn't make it longer tho}

\begin{figure}[!t]
    \centering

    \includegraphics[
        width=0.94\textwidth,
        keepaspectratio
    ]{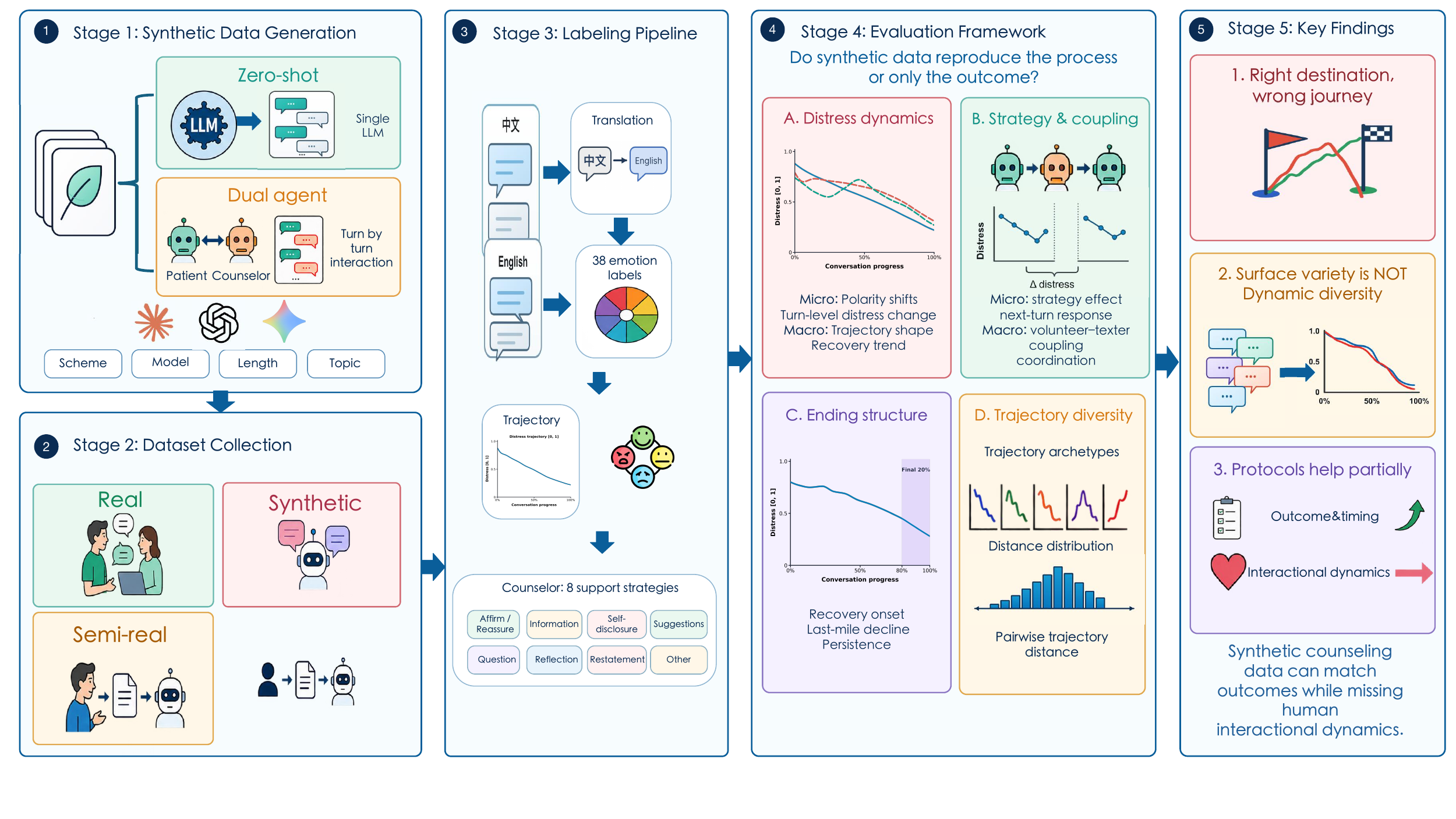}

    \captionsetup{skip=4pt}
    \caption{
   \textbf{ Overview }of our framework for evaluating process fidelity in synthetic
    counseling dialogues. We compare real, semi-real, and synthetic datasets across various generation protocols and then evaluate process fidelity across emotional trajectories,
    interactional responsiveness, recovery structure, and trajectory-level
    diversity. We use the resulting failure patterns to guide improvements
    to synthetic data generation.
    }
    \label{fig:main}
\end{figure}

\section{Datasets}
\label{sec:datasets}
\begin{table}[t]
\centering
\small
\renewcommand{\arraystretch}{1.10}
\setlength{\tabcolsep}{5pt}

\begin{tabularx}{\textwidth}{
@{}
l
>{\raggedright\arraybackslash}X
r
r
@{}}

\toprule
\textbf{Corpus}
& \textbf{Description}
& \textbf{\# Conv.}
& \textbf{Avg.\ Utts.} \\
\midrule

\rowcolor{gray!15}
\multicolumn{4}{@{}l}{\textbf{\textit{Real}}} \\
ESConv~\citep{liu-etal-2021-towards}
& Peer-support chats; ESC framework
& 1,300
& 29.5 \\

\midrule
\rowcolor{gray!15}
\multicolumn{4}{@{}l}{\textbf{\textit{Semi-real}}} \\
AnnoMI~\citep{wu2023annomi}
& Performed MI dialogues
& 133
& 72.9 \\

PsyDial~\citep{qiu-lan-2025-psydial}
& Real counselor turns, regenerated patients
& 2,382
& 75.6 \\

\midrule
\rowcolor{gray!15}
\multicolumn{4}{@{}l}{\textbf{\textit{Synthetic}}} \\
CACTUS~\citep{lee-etal-2024-cactus}
& GPT-4o; intake forms + CBT plans
& 2,173
& 30.5 \\

PsychEval~\citep{pan-etal-2026-psycheval}
& GPT-5; clinical cases; multi-session therapy
& 2,798
& 49.1 \\

SimPsyDial~\citep{qiu-lan-2026-interactive}
& GPT-4; grounded in real help-seeking posts
& 1,000
& 25.9 \\

\textcolor{revisionblue}{PsyChain~\citep{feng-etal-2026-psychain}} &
\textcolor{revisionblue}{Collaborative chain-of-agents generation} &
\textcolor{revisionblue}{1,047} &
\textcolor{revisionblue}{37.0} \\

\midrule
\rowcolor{gray!15}
\multicolumn{4}{@{}l}{\textbf{\textit{Controlled conditions(Ours)}}} \\
\textbf{Baseline}
& Generation without topic or length constraints
& 660
& 47.6 \\

\textbf{Dual Agent}
& Separate patient and counselor agents
& 190
& 46.2 \\

\textbf{Zero-shot}
& Single-pass model generation
& 1,200
& 86.6 \\

\bottomrule
\end{tabularx}

\caption{
Overview of the counseling-dialogue datasets used in this study.
Average utterance counts include turns from both participants. Detailed dataset
descriptions are provided in
Appendix~\ref{app:dataset_details}.
}
\label{tab:datasets}

\end{table}
%\sara{add references to teh datasets here; maybe also make the table to compact...}
We evaluate \textcolor{revisionblue}{ten} counseling-dialogue datasets spanning real, semi-real,
and fully synthetic interactions (Table~\ref{tab:datasets}). We select
existing open-source datasets to cover different levels of realism and representative
generation paradigms:

\textbf{Real.} ESConv is our primary human--human reference because
    it is publicly accessible and contains English conversations between
    help-seekers and trained peer supporters.\\
\textbf{Semi-real. } AnnoMI contains English, professionally performed
    motivational-interviewing demonstrations, while PsyDial is a Chinese-language
    dataset that preserves real counselor turns and regenerates the patient side
    with LLMs.\\
\textbf{Synthetic. } CACTUS generates complete conversations from
    intake forms and CBT plans using a single model; PsychEval is a
    Chinese-language, multi-session dataset generated from clinical case reports
    across multiple therapy styles; and SimPsyDial uses separately prompted
    counselor and patient agents grounded in real help-seeking posts. \textcolor{revisionblue}{PsyChain employs a chain-of-agents framework to construct conseling dialogues through structured simulation.}
% \begin{itemize}
%     \item \textbf{Real.} ESConv is our primary human--human reference because
%     it is publicly accessible and contains English conversations between
%     help-seekers and trained peer supporters.

%     \item \textbf{Semi-real.} AnnoMI contains English, professionally performed
%     motivational-interviewing demonstrations, while PsyDial is a Chinese-language
%     dataset that preserves real counselor turns and regenerates the patient side
%     with LLMs.

%     \item \textbf{Synthetic.} CACTUS generates complete conversations from
%     intake forms and CBT plans using a single model; PsychEval is a
%     Chinese-language, multi-session dataset generated from clinical case reports
%     across multiple therapy styles; and SimPsyDial uses separately prompted
%     counselor and patient agents grounded in real help-seeking posts.
% \end{itemize}

Existing synthetic datasets differ simultaneously in prompts, models,
grounding, counseling protocols, and generation procedures, making it difficult
to isolate which design choices produce particular process patterns. Recent work has explored LLM-generated counseling interactions to address the
scarcity and access restrictions of authentic clinical dialogue, including
role-simulated patients for counselor training
\citep{louie-etal-2024-roleplay,louie2026llm}
and synthetic psychotherapy dialogues for privacy-preserving development of
clinical NLP tools \citep{cabrera-synthetic}.
Therefore, we adapt the patient-simulation prompts from prior work
\citep{louie-etal-2024-roleplay,louie2026llm} and use them to construct three
controlled synthetic-generation conditions: \textbf{Baseline},
\textbf{Zero-shot}, and \textbf{Dual Agent}. As the original prompts are
designed primarily for the patient role, we construct a corresponding
counselor prompt by mirroring the patient-side structure and adapting its role
instructions to the counselor. Baseline contains our initial generation runs without explicit topic or length constraints; in subsequent analyses, we separate it into Baseline ZS (zero-shot) and Baseline DA (dual-agent) by generation mode. \textbf{Zero-shot} and \textbf{Dual Agent} settings add the corresponding controlled generation settings described in Appendix~\ref{sec:generation_conditions}. Dataset preprocessing and labeling follow \empath{}~\citep{gong2026empathtracingmultilevelemotion}.

% Therefore, we add three controlled conditions: \textbf{Baseline}, \textbf{Zero-shot}, and \textbf{Dual Agent}, which vary by
% generation mode under ablation settings. In subsequent analyses, we further
% separate the Baseline condition by generation mode into
% \textbf{Baseline ZS} (zero-shot) and \textbf{Baseline DA} (dual-agent).
% Dataset construction and preprocessing follow
% \empath{}~\citep{gong2026empath}, with additional details in
% Appendix~\ref{app:dataset_details}.
% \msnote{This section is much better! The appendix ref at the end doesn't seem to go to the right place. And is the dataset construction and preprocessing for all datasets or just synthetic data? It makes it sound liek EMPATH also proposed the synthetic data generation parts which was not my understanding and changes the contribution here a little. Lastly, can you add one sentence giving us a preview of what baseline is? It's not intuitive how baseline zero-shot would be different from zero-shot.}

%\msnote{I think this section could be condensed and would read easier with some more structure (e.g., bulleted or paragraphed list of datasets in the same order as the summary table, with a couple sentences about each. Also can you motivate what your synthetic datasets are meant to achieve that the existing synthetic datasets do not? Also, you should specify which datasets are in Chinese.}

\section{Methods}
\label{sec:methods}

As summarized in Figure~\ref{fig:main}, we evaluate whether synthetic
counseling dialogues reproduce the \emph{process} of real, authentic conversations
rather than only their outcomes. 
%\ndnote{Probably need a "building on citet{arxiv-placeholder}" here or something. If we want to try to hide the overlap in authors, could probably cite other emotion dynamics work here as well.}
We first label all datasets with utterance-level emotion labels and counselor
support-strategy labels (Stage~3), and then compute distress scores for all
patient utterances. We next evaluate four aspects of process fidelity
(Stage~4): \textbf{(A) Emotion and Distress Dynamics},
\textbf{(B) Strategy and Coupling}, \textbf{(C) Recovery Structure}, and
\textbf{(D) Trajectory Diversity}. Together, these analyses move from local
turn-level changes to whole-conversation trajectories and corpus-level
behavioral coverage. Finally, we use the observed gaps to guide targeted
changes to synthetic generation and test whether process fidelity can
be improved.
%\sara{As we cut down, we can mainly focus on this section by referring directly to the arxiv of the CTL paper. Let's use "arxiv-placeholder" for now, and I'll update the citatations once the official arxiv is up. i.e. use \citep{arxiv-placeholder} whenever you need to reference}

\subsection{Emotion Labeling}
\label{sec:annotation}

%sara{thinking about better wording than "annotation".. }
%\ndnote{We just had a discussion where someone said that ICLR doesn't like the word "pipeline". Not sure how much stock to put in that, but maybe just say "approach"?} 

% \lareina{changed title to emotion labeling, but not sure if appropriate}

We build our labeling approach following \empath{} \citep{gong2026empathtracingmultilevelemotion}, which provides the emotion taxonomy, affective mappings, and labeling framework used in our analysis. \empath{} uses a Llama model locally for emotion labeling due to privacy constraints on crisis-counseling data. For publicly available datasets and our generated synthetic datasets, we instead use GPT-5.6-luna~\citep{openai_gpt56_luna} as a more efficient alternative with strong performance on this task. \textcolor{revisionblue}{
To assess annotation reliability, we compare GPT-5.6 Luna,
Claude Opus 4.8, and Gemini 3.6 Flash on 705 client turns
from 50 ESConv conversations, observing substantial
pairwise emotion-label agreement (Cohen's
$\kappa=0.62$--$0.68$) and strong distress-score
correlations ($r=0.81$--$0.85$). 
We select GPT-5.6 Luna for the full annotation pipeline
because it achieves comparable cross-model consistency
at substantially lower inference cost (approximately
20$\times$ cheaper than Opus 4.8 and 3$\times$ cheaper
than Gemini 3.6 Flash). More details about model selection is provided in Appendix~\ref{app:emotion_annotation_reliability}.
}We retain the \empath{} taxonomy with one refinement: the original \textit{self} category is split into \textit{self-doubt} and \textit{self-awareness}, yielding 38 emotion labels in total. For Chinese-language datasets, we first translate utterances into English using GPT-5.4-mini~\citep{openai_gpt54mini}. We then apply the same labeling approach to all datasets. Data preprocessing and emotion-to-valence and distress mappings follow \empath{}.
% The data
% pre-processing method,
% \empath{}. Emotion-to-valence and distress mappings, as well as the
% eight-category counselor support-strategy labeling, all follow
% \empath{}.

\subsection{Analysis Framework}
% \textbf{A. Emotion and Distress Dynamics.}
\paragraph{\textbf{A. Emotion and Distress Dynamics.}}
\label{sec:methods_micro}
\textit{At turn level}, we follow the micro-analysis setup of \empath{} ~\citep{gong2026empathtracingmultilevelemotion} to
characterize local emotional change over consecutive patient turns. \textbf{SCR} denotes the mean per-conversation sentiment change rate,
capturing how frequently patient sentiment changes across adjacent turns;
\textbf{Stable} is the percentage of conversations with no sentiment change;
\textbf{Turn} is the median turn of the first negative-to-positive transition
among conversations containing one; \textbf{Ratio} is the ratio of recovery
(\textit{neg-neg-pos}) to relapse (\textit{neg-pos-neg}) trigrams; and
\textbf{Rec} is the percentage of overlapping trigrams exhibiting the recovery
pattern. We also report the proportions of conversations that begin negative
and end positive. SCR rankings are robust to the valence cut-points (Spearman $\rho\geq0.86$ under
0.35/0.65, a binary split, or dropping neutral turns), though the real--synthetic
gap narrows without a neutral class. \textit{At the trajectory level}, we use distress scores defined in
\empath{} to characterize whole-conversation emotional
dynamics. Each patient trajectory is linearly resampled onto 20 equally
spaced points of normalized conversation progress and averaged pointwise across
conversations with 95\% normal-approximation confidence bands. Unlike the
turn-level analysis which is sensitive to starting emotion, the trajectory analysis uses untrimmed sequences and
retains conversations with at least four patient turns, so corpus counts may
differ slightly across analyses. \textcolor{revisionblue}{
In addition, we assess sensitivity to interpolation resolution
using 30, 40, and 50 points (Appendix~\ref{app:interpolation}). Turn-level metrics are computed directly from the
original sequences and are therefore unaffected
by interpolation resolution.}

%\ndnote{(See higher-level comment below) Minor and maybe can just be done in preparation for rebuttals, but it may be good to check how sensitive the results are to the valence thresholds from earlier; i.e., if there is a corpus where conversation turns hover between .4-.6, then the SCR will be high potentially due to noise. I might just see whether the trends in SCR change significantly if you use like .65 and .35 or drop 0 entirely and just map to - and + 1.}

% \textbf{B. Strategy and Coupling.}
\paragraph{\textbf{B. Strategy and Coupling.}}
\label{sec:strategy}
To test whether patient emotion change is associated with counselor behavior, we analyze three-turn windows $(u_t,c_t,u_{t+1})$ consisting of a patient turn, a counselor turn, and the subsequent patient turn. For each counselor strategy, we report its frequency and the subsequent distress change $\Delta d_t=d(u_{t+1})-d(u_t)$. We further quantify how informative counselor behavior is about the patient's next emotional state using conditional mutual information (CMI). Let $X_t$ denote the counselor's strategy (8 ESConv categories) or emotion, and $Y_t$ and $Y_{t+1}$ the patient's current and subsequent emotional states, with emotional states discretized into low, medium and high distress. We compute $I(X_t;Y_{t+1}\mid Y_t)$, which measures the additional information about the patient's next state provided by counselor behavior beyond the patient's current state. Because the plug-in estimator is biased upward, we subtract the mean of a permutation null in which counselor turns are shuffled among windows with the same current patient state, preserving the patient's own transition structure. For the stage-adjusted estimate, we additionally restrict these permutations to turns in the same fifth of the conversation, thereby removing apparent coupling that can arise when both speakers independently follow the same conversation-level progression. Confidence intervals come from a bootstrap over conversations. Additional details are provided in Appendix~\ref{app:volunteer_coupling}.

% \textbf{C. Recovery Structure.}
\paragraph{\textbf{C. Recovery Structure.}}
\label{sec:recovery}
% \ndnote{(See higher-level comment below) Similar here, would wonder if results are the same with a different threshold.}
%\ndnote{I would also rephrase to specify that we are defining recovery as dt <= .33; currently its phrased like this was already defined earlier or in prior work.}

We characterize whole-conversation recovery using three complementary
statistics. We define a patient turn as \emph{recovered} when its distress
score satisfies $d_t\leq0.33$. We set \(d_t\le0.33\) because it isolates explicitly positive/low-distress states while excluding neutral and mixed-distress states. With our label-to-distress mapping, this
admits the explicitly positive states (e.g., \emph{hopeful}, \emph{gratitude},
\emph{serenity}) and \emph{anticipation}, but not \emph{self-aware},
\emph{longing}, or \emph{neutral}. \emph{Recovery onset} is the normalized
position of the first recovered turn. \emph{Last-mile share} measures the
fraction of net distress reduction occurring in the final 20\% of turns to study the final portion of the conversation more closely.
\emph{Persistence} is the share of post-onset turns that remain recovered.
Together, these measures capture when recovery begins, how it accumulates,
and whether it holds once reached. 

%\ndnote{(Higher-level commetn) I feel like the metrics in recovery structure in particular as well as the thresholds chosen for some measures feel like they could use more motivation for each. E.g., why do we necessarily care about the fraction of net distress reduction in the last 20\% of turns? And why the lat 20\% in particular? I get there probably isn't room to elaborate a ton, but at the moment it feels a bit easy for a reviewer to push back against the measures we chose and ask us to try other thresholds.}

\begin{table}[t]
\centering
\small
\setlength{\tabcolsep}{3.5pt}
\resizebox{\textwidth}{!}{%
\begin{tabular}{c
                l r l r l r
                l r l r l r}
\toprule
& \multicolumn{6}{c}{\textbf{patient}}
& \multicolumn{6}{c}{\textbf{counselor}} \\
\cmidrule(lr){2-7}
\cmidrule(lr){8-13}
& \multicolumn{2}{c}{\textbf{Real}}
& \multicolumn{2}{c}{\textbf{Semi-real}}
& \multicolumn{2}{c}{\textbf{Synthetic}}
& \multicolumn{2}{c}{\textbf{Real}}
& \multicolumn{2}{c}{\textbf{Semi-real}}
& \multicolumn{2}{c}{\textbf{Synthetic}} \\
\cmidrule(lr){2-3}
\cmidrule(lr){4-5}
\cmidrule(lr){6-7}
\cmidrule(lr){8-9}
\cmidrule(lr){10-11}
\cmidrule(lr){12-13}
\textbf{Rank}
& \textbf{Label} & \textbf{\%}
& \textbf{Label} & \textbf{\%}
& \textbf{Label} & \textbf{\%}
& \textbf{Label} & \textbf{\%}
& \textbf{Label} & \textbf{\%}
& \textbf{Label} & \textbf{\%} \\
\midrule

1 & neutral & 16.91
  & neutral & 25.17
  & self-doubt & \textcolor{revisionblue}{13.98}
  & neutral & 61.52
  & neutral & 73.62
  & neutral & \textcolor{revisionblue}{54.51} \\

2 & gratitude & 9.55
  & self-doubt & 8.91
  & fear & \textcolor{revisionblue}{10.22}
  & empathy & 13.25
  & empathy & 10.09
  & empathy & \textcolor{revisionblue}{24.11} \\

3 & hopeful & 7.68
  & self-aware & 8.45
  & anxiety & \textcolor{revisionblue}{8.38}
  & hopeful & 11.21
  & hopeful & 3.08
  & hopeful & \textcolor{revisionblue}{8.76} \\

4 & stress & 7.43
  & hopeful & 6.08
  & \textcolor{revisionblue}{self-aware}
  & \textcolor{revisionblue}{6.83}
  & happiness & 1.73
  & distress & 1.36
  & \textcolor{revisionblue}{fear}
  & \textcolor{revisionblue}{1.62} \\

5 & disappointment & 6.71
  & disappointment & 5.54
  & \textcolor{revisionblue}{hopeful}
  & \textcolor{revisionblue}{6.71}
  & stress & 1.51
  & worry & 1.32
  & \textcolor{revisionblue}{self-aware}
  & \textcolor{revisionblue}{1.47} \\

\bottomrule
\end{tabular}%
}
\caption{
Top five emotion labels for patient and counselor turns,
pooled by data tier. \textcolor{revisionblue}{PsyChain included and numbers updated in blue.}
}
\label{tab:emotion_distribution}
\end{table}

% \textbf{D. Trajectory Diversity.}
\paragraph{\textbf{D. Trajectory Diversity.}}
\label{sec:diversity}
%\ndnote{May want to specify like "Trajectory Diversity" to not make readers think this is like an n-gram diversity measure}
Finally, we evaluate whether these datasets cover diverse emotional trajectories in their conversations
or repeatedly generates variants of the same underlying pattern. We
interpolate each patient distress trajectory onto a common 20-point grid and
cluster pooled trajectories with $k$-means ($k=5$) to obtain shared recovery
archetypes. We compare each corpus's distribution over these archetypes and
its mean pairwise trajectory distance. For comparison, we also compute
pairwise Jensen--Shannon divergence over per-conversation emotion and
strategy distributions, separating temporal trajectory diversity from
surface-level label diversity. 
% \lareina{smoothing?} \sara{no need to say smoothing}

\section{Experiments and Results}
%\subsection{Micro-analysis} \lareina{maybe combine with 4.4.1 since they are essentially talking about the same thing, and I in the methods section micro is merged into macro} \ndnote{If we get desperate for space, Kind of wonder if it can just be dropped, unless I'm mis-grouping something, I feel like the most interesting results come out of the macro-analysis parts. If can't drop, I do agree about merging though.}
\label{sec:micro}

\subsection{Do synthetic patients recover like real Ones?}
\label{sec:micro}

% \vspace{-0.6em}
\begin{wrapfigure}{t}{0.65\textwidth}
    % \vspace{-0.8em}
    \centering
    \includegraphics[width=\linewidth]{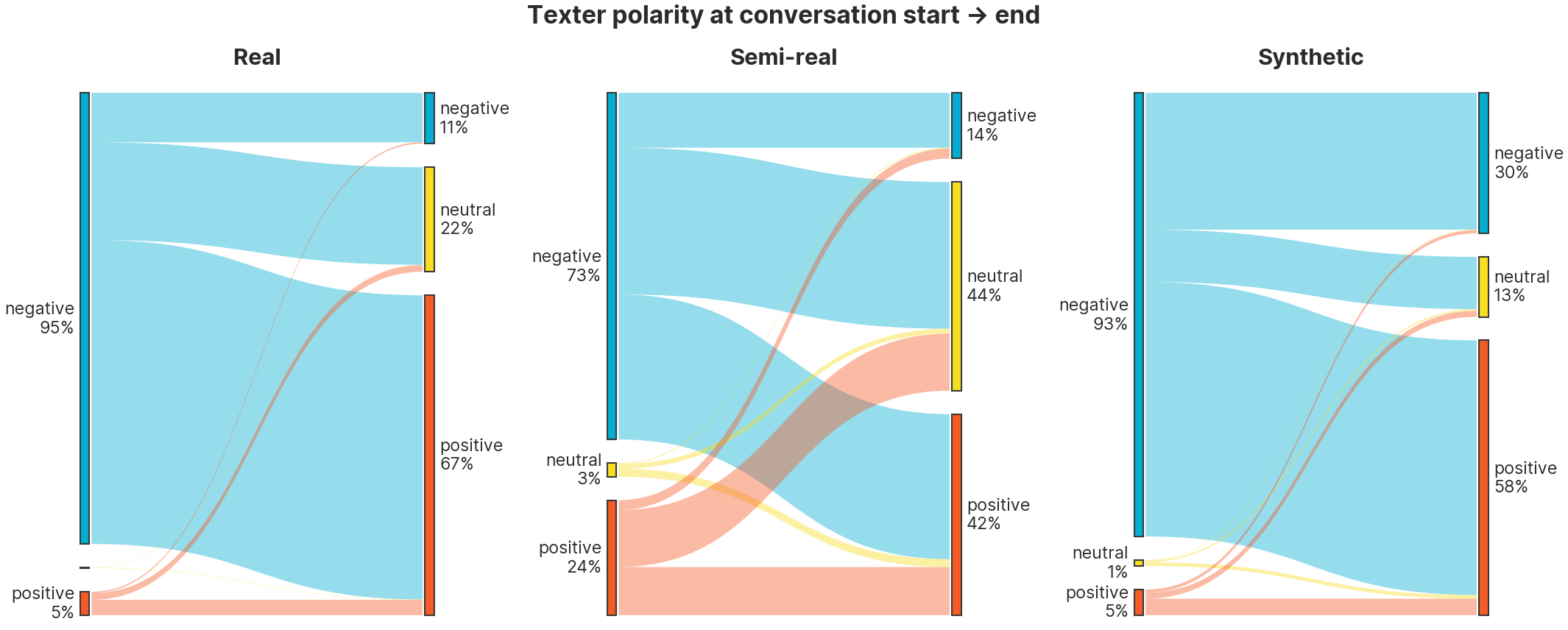}
    \captionsetup{font=footnotesize,skip=2pt}
    \caption{
    Patient sentiment at conversation start and end, averaged unweighted
    over datasets within each data tier. 
    % \textcolor{revisionblue}{PsyChain included.}
    }
    \label{fig:polarity_sankey}
    %\vspace{-0.8em}
\end{wrapfigure}

%\ndnote{Since I imagine some will just scan for these headings, narrower than what?} 
\textbf{Synthetic datasets show a narrower emotional range than authentic conversations.} As shown in Table~\ref{tab:emotion_distribution}, patient turns in real
conversations are distributed across neutral, positive, and negative emotion
labels, whereas synthetic patient turns are concentrated in a smaller set of
distress-related states, particularly \textit{self-doubt}, \textit{fear}, and
\textit{anxiety}. This narrower range is also reflected in emotion transitions, as detailed in Appendix Table~\ref{tab:top_emotion_transitions}. In our three generated
datasets, the five most frequent transitions are dominated by repetitions of
negative states. CACTUS and SimPsyDial additionally include a direct
\textit{self-doubt}$\rightarrow$\textit{hopeful} transition among their most
frequent patterns, suggesting a comparatively abrupt shift toward recovery.
By contrast, ESConv shows a more varied transition profile, including
persistence in neutral states and \textit{gratitude}.

\begin{figure}[t]
    \centering
    \captionsetup{skip=3pt}

    % ===================== Row 1 =====================
    \begin{minipage}[t]{0.48\textwidth}
        \centering
        \includegraphics[width=\linewidth]{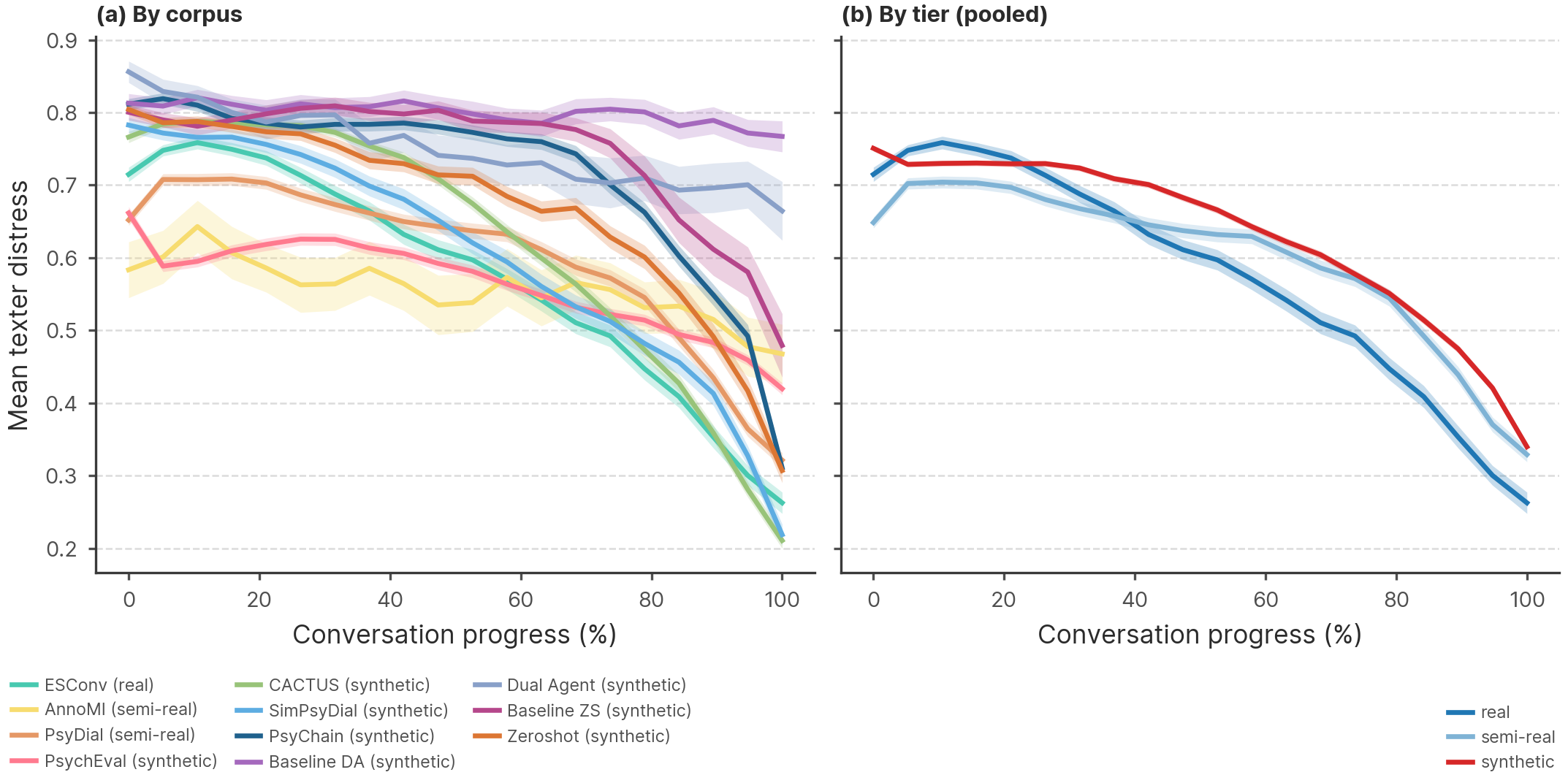}

        {\scriptsize (a) Distress trajectories}
    \end{minipage}
    \hfill
    \begin{minipage}[t]{0.48\textwidth}
        \centering
        \includegraphics[width=\linewidth]{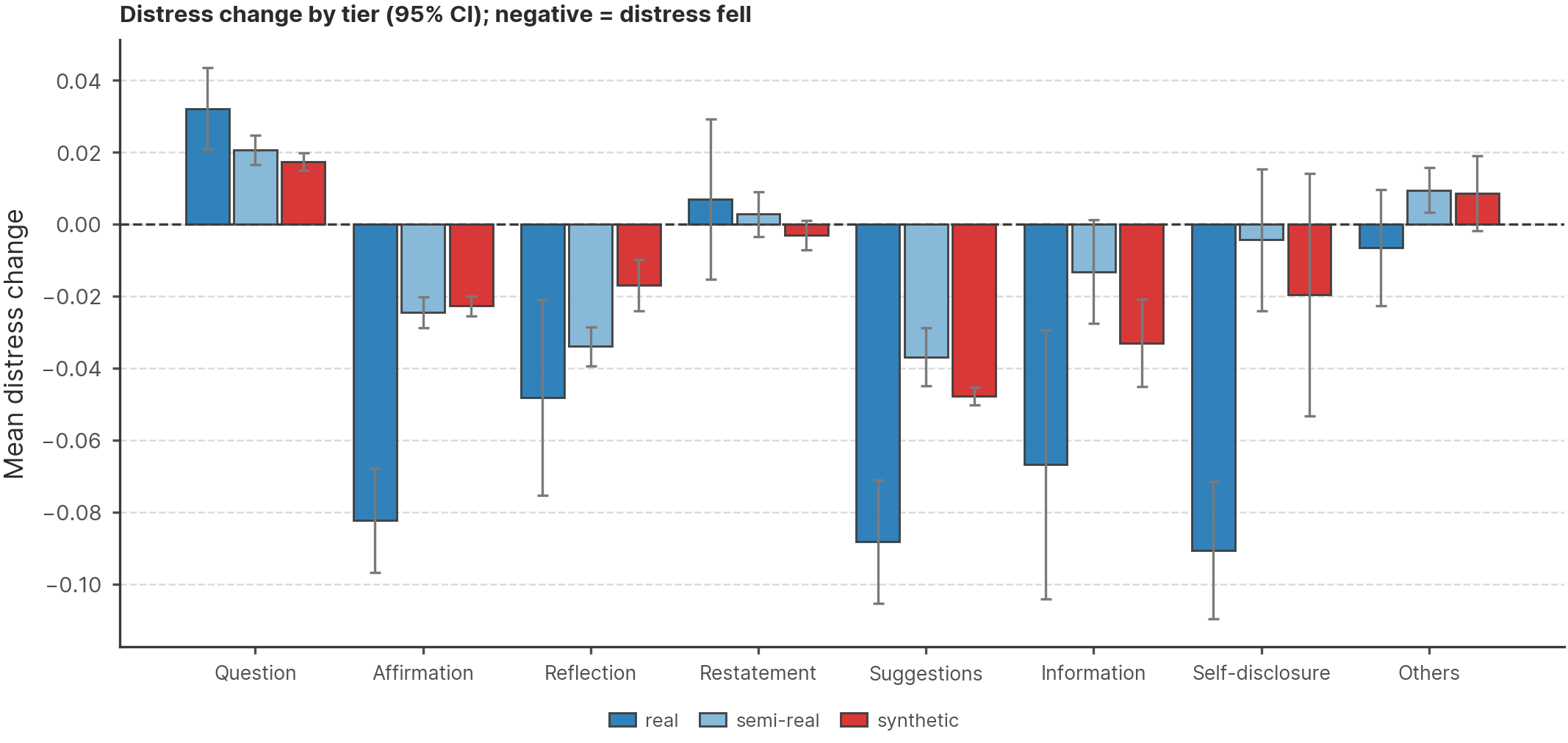}

        {\scriptsize (b) Strategy downstream effect}
    \end{minipage}

    %\vspace{0.3em}

    % ===================== Row 2 =====================
    \begin{minipage}[t]{0.48\textwidth}
        \centering
        \includegraphics[width=\linewidth]{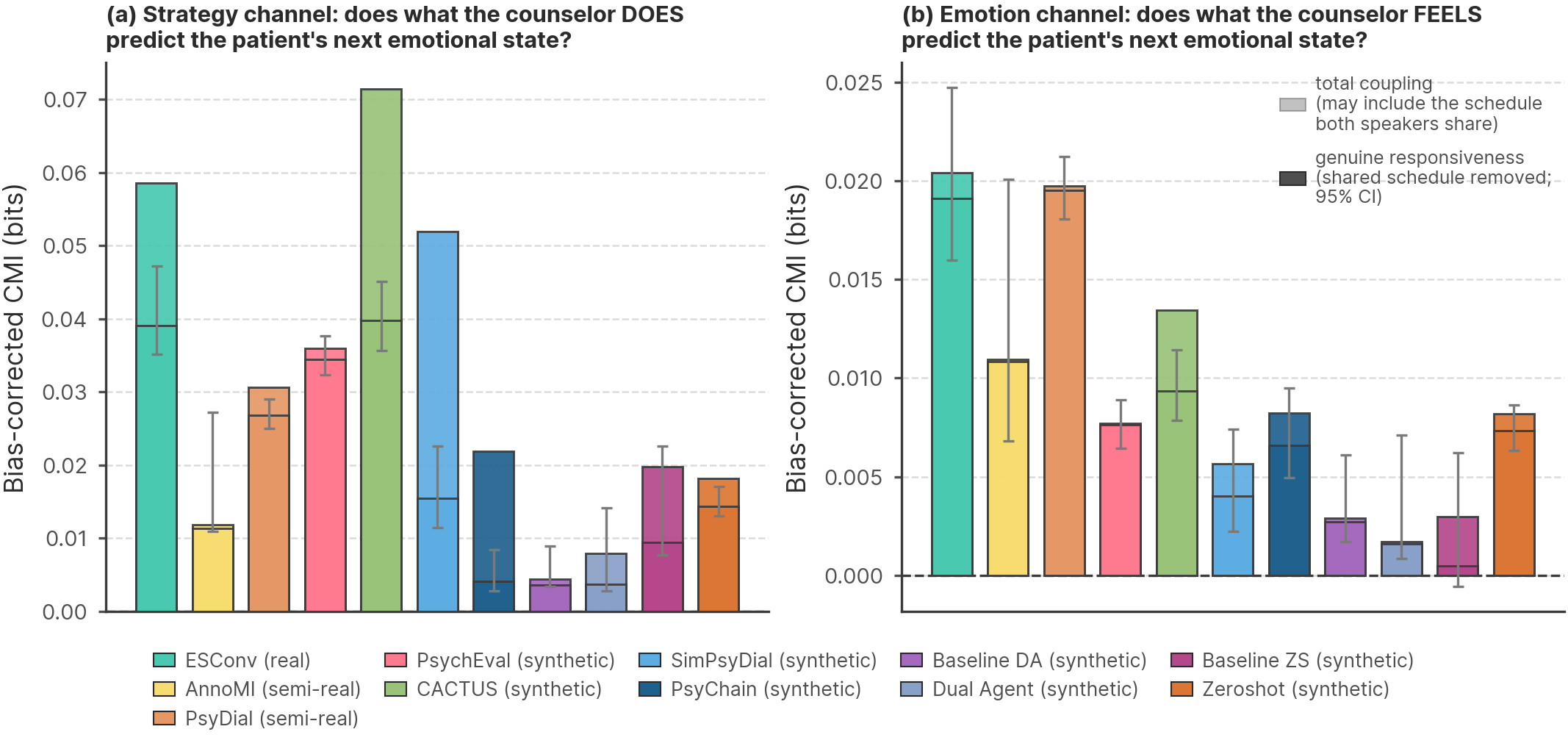}

        {\scriptsize (c) Strategy--emotional coupling}
    \end{minipage}
    \hfill
    \begin{minipage}[t]{0.48\textwidth}
        \centering
        \includegraphics[width=\linewidth]{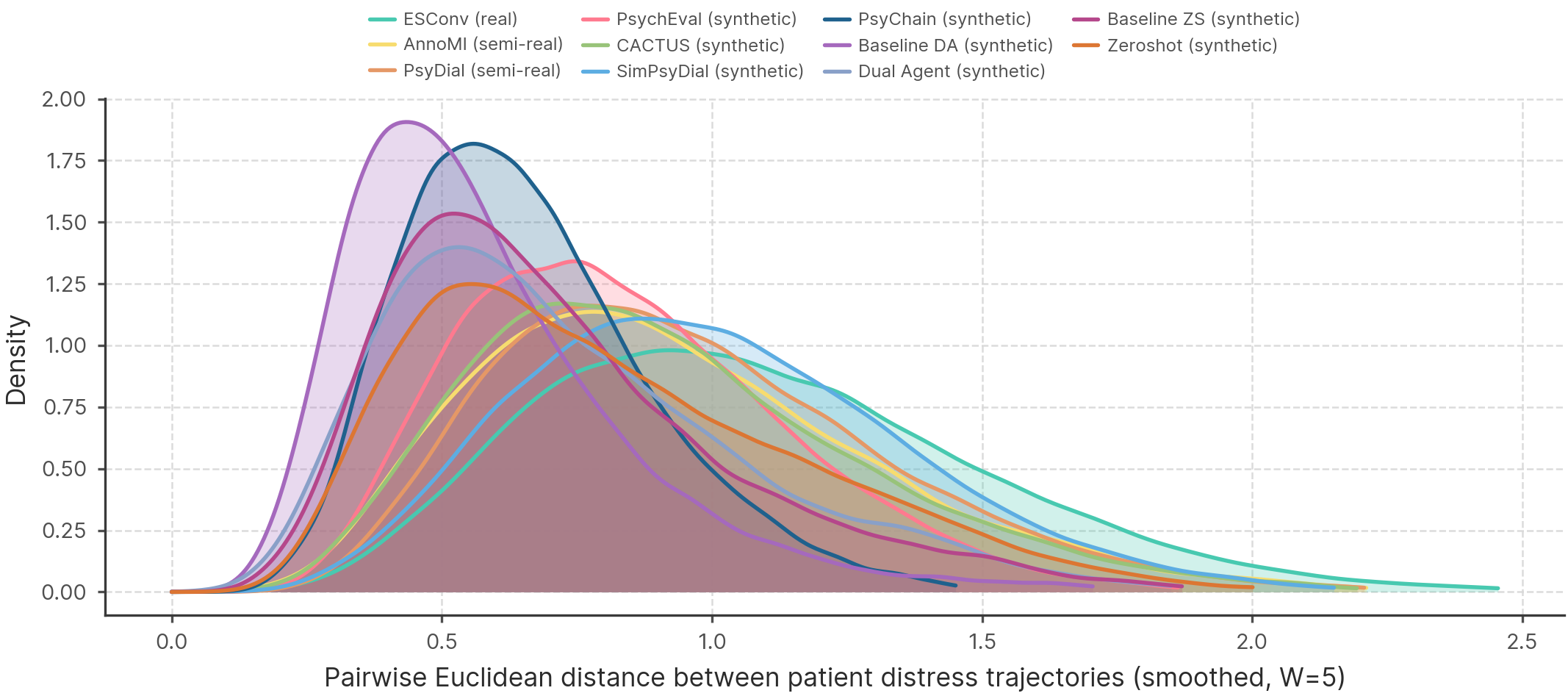}

        {\scriptsize (d) Inter-conversation diversity}
    \end{minipage}

    \caption{
    Comparison of process-level properties across real and synthetic
    counseling dialogues. \textcolor{revisionblue}{PsyChain included.}
    %\msnote{Try to increase the font size in these figures}
    }
    \label{fig:process_comparison}
\end{figure}

\textbf{Synthetic dialogues can reach plausible endpoints while following
different recovery dynamics from authentic conversations.}
In real ESConv conversations, the Sankey flow (Figure~\ref{fig:polarity_sankey}) shows that
95.2\% of conversations begin negative but 67.3\% end positive, despite
frequent sentiment shifts (SCR$=0.386$, see Table~\ref{tab:micro_summary}). At the continuous level, distress declines gradually after early problem exploration, as shown in Figure~\ref{fig:process_comparison}(a). This indicates that in real conversations,
recovery begins around mid-conversation (onset $\sim 0.56$), accumulates throughout
the dialogue rather than only near the end (last-mile share $\sim 0.49$), and persists
for roughly half of subsequent turns (see Figure~\ref{fig:scripted_recovery}). 
\textcolor{revisionblue}{
All nine comparison corpora exhibit significantly
different distress trajectories from ESConv, both
in overall RMS distance and after mean-centering
to isolate trajectory shape (Holm-adjusted
$p<0.001$; Appendix~\ref{app:significance_trajectory}).
}

\begin{wraptable}{H}{0.64\textwidth}
\centering
\scriptsize
\setlength{\tabcolsep}{3pt}
\renewcommand{\arraystretch}{1.05}

\begin{tabular}{@{}lrrrrrrrr@{}}
\toprule
& &
\multicolumn{1}{c}{\textbf{Start}} &
\multicolumn{2}{c}{\textbf{Volatility}} &
\multicolumn{1}{c}{\textbf{End}} &
\multicolumn{3}{c}{\textbf{Recovery signature}} \\
\cmidrule(lr){3-3}
\cmidrule(lr){4-5}
\cmidrule(lr){6-6}
\cmidrule(lr){7-9}

\textbf{Corpus} &
\textbf{$N$} &
\textbf{Neg\,\%} &
\textbf{SCR} &
\textbf{Stable\,\%} &
\textbf{Pos\,\%} &
\textbf{Turn} &
\textbf{Ratio} &
\textbf{Rec\,\%} \\
\midrule

\rowcolor{gray!15}
\multicolumn{9}{@{}l}{\textbf{\textit{Real}}} \\
ESConv
& 1,300 & 95.2 & 0.386 & 0.8 & 67.3
& 6.0 & 1.95 & 7.8 \\

\midrule
\rowcolor{gray!15}
\multicolumn{9}{@{}l}{\textbf{\textit{Semi-real}}} \\
AnnoMI
& 130 & 63.8\sigthree & 0.407
& 5.4\sigtwo & 27.7\sigthree
& 8.0\sigtwo & 2.00 & 3.0\sigthree \\

PsyDial
& 2,382 & 83.6\sigthree & 0.335\sigthree
& 0.6 & 56.9\sigthree
& 8.0\sigthree & 1.44\sigthree & 7.9 \\

\midrule
\rowcolor{gray!15}
\multicolumn{9}{@{}l}{\textbf{\textit{Synthetic}}} \\
Baseline DA
& 360 & 99.7\sigthree & 0.064\sigthree
& 56.9\sigthree & 7.2\sigthree
& 14.0\sigthree & 1.28\sigthree & 3.6\sigthree \\

Baseline ZS
& 300 & 100.0\sigthree & 0.135\sigthree
& 17.3\sigthree & 48.3\sigthree
& 14.0\sigthree & 1.77 & 6.0\sigthree \\

CACTUS
& 2,173 & 95.3 & 0.211\sigthree
& 4.0\sigthree & 78.2\sigthree
& 9.0\sigthree & 2.01 & 9.3\sigthree \\

Dual Agent
& 190 & 97.4 & 0.121\sigthree
& 43.7\sigthree & 19.5\sigthree
& 11.0\sigthree & 1.26\sigthree & 7.1 \\

PsychEval
& 2,798 & 73.3\sigthree & 0.508\sigthree
& 0.0\sigthree & 48.0\sigthree
& 6.0 & 1.43\sigthree & 4.9\sigthree \\

SimPsyDial
& 1,000 & 92.4\sigone & 0.330\sigthree
& 1.5 & 84.2\sigthree
& 5.0 & 1.85 & 13.7\sigthree \\

\textcolor{revisionblue}{PsyChain}
& \textcolor{revisionblue}{1,047}
& \textcolor{revisionblue}{96.8}
& \textcolor{revisionblue}{0.220\sigthree}
& \textcolor{revisionblue}{0.3}
& \textcolor{revisionblue}{88.2\sigthree}
& \textcolor{revisionblue}{11.0\sigthree}
& \textcolor{revisionblue}{1.65\sigthree}
& \textcolor{revisionblue}{9.8\sigthree} \\

Zero-shot
& 1,200 & 98.3\sigthree & 0.242\sigthree
& 0.4 & 60.8\sigthree
& 16.0\sigthree & 1.33\sigthree & 7.7 \\

\bottomrule
\end{tabular}

\caption{
Patient-level sentiment dynamics across datasets.
\textcolor{revisionblue}{
Superscripts denote Holm-adjusted significance
relative to ESConv across 10 corpus comparisons
per metric:
$^{\dagger}p_{\mathrm{Holm}}<0.05$,
$^{\ddagger}p_{\mathrm{Holm}}<0.01$,
and $^{\S}p_{\mathrm{Holm}}<0.001$.
}}
\label{tab:micro_summary}
\end{wraptable}

Synthetic datasets, however, diverge from this pattern. Some remain
stuck in distress, while others reach positive endings through qualitatively
different recovery paths. Baseline DA and Dual Agent show \emph{plateau
distress}, with 56.9\% and 43.7\% of conversations remaining in the same sentiment polarity throughout. 
% showing little change in patient sentiment polarity across turns. 
CACTUS and SimPsyDial show more \emph{scheduled recovery}, consistent
with generation toward a predefined reframed thought or termination at
farewell/encouragement cues. SimPsyDial, for
example, ends positive in 84.2\% of conversations but exhibits substantially
more recovery trigrams than ESConv (13.7\% vs.\ 7.8\%) and an earlier first
negative-to-positive transition (5.0 vs.\ 6.0 turns). Zero-shot shows more
\emph{back-loaded recovery}, with later negative-to-positive transitions and
improvement concentrated near the end, while PsychEval exhibits
\emph{mandated churn}: its prompts explicitly encourage non-linear emotion,
resistance, and topic shifts, yielding the highest
sentiment change rate (SCR$=0.508$) without a correspondingly high
positive-ending rate. \textcolor{revisionblue}{These differences are statistically significant for most sentiment-dynamics metrics (Holm-adjusted \(p < 0.05\); Appendix~\ref{tab:micro_summary_significance}). In particular, all synthetic corpora differ significantly from ESConv in sentiment change rate (SCR) and positive-ending rate (both \(p_{\mathrm{Holm}} < 0.001\)).}
% Overall, matching the final state does not imply matching the temporal process through which recovery unfolds.

Overall, synthetic data generation methods often achieve
the trajectory properties they explicitly target, such
as positive endings or staged emotional change, but
these targets do not ensure that recovery unfolds as it
does in authentic conversations. Matching the final
state, or even prescribing parts of the trajectory,
does not imply reproducing the interactional process
through which recovery emerges.

\textcolor{revisionblue}{
Across all eight synthetic corpora,
we find significant differences from ESConv in distress
trajectory shape, sentiment change rate
and positive-ending rate, as well as
consistently lower inter-conversation trajectory
diversity and weaker emotion-channel
coupling, all with Holm-adjusted
$p<0.01$. However, the direction of sentiment
differences and the patterns of recovery timing,
strategy use, and strategy-channel coupling vary
across datasets and are subject to generation paradigms. These findings suggest that certain
process-level discrepancies persist across generation
framework, while others depend on the specific
generation approach.
}

This raises a natural question: can process fidelity
be improved simply by specifying these missing
details more explicitly? We test this directly in
Section~\ref{sec:scorecard}.

\subsection{Do synthetic counselors support patients like real counselors?}
\label{sec:strategy}

\textbf{Synthetic datasets contain less diverse counselor behavior, with weaker downstream effects and weaker coupling to patient emotion.}
Table~\ref{tab:strategy_summary} and Figure~\ref{fig:strategy_pairs} in the appendix show that in ESConv, counselor strategies are relatively evenly distributed, and no single strategy accounts for more than 24\% of turns. 
In every synthetic dataset, by contrast, counselor behavior is concentrated in the two most frequent strategies. Behaviors that commonly appear in real peer support, such as \textit{self-disclosure, greetings, small talk, and logistics}, also rarely appear in synthetic datasets.
The observed gap in strategy distributions also correspond closely to the counselor prompts used for generation: PsychEval and CACTUS are question- and suggestion-heavy, whereas our three controlled conditions converge on affirmation under the same validation-focused prompt. However, this narrower repertoire does not translate into stronger downstream effects on patient distress. In ESConv, self-disclosure, suggestions, and affirmation are associated with the largest immediate reductions in distress, as shown in Figure~\ref{fig:process_comparison}(b). In semi-real and synthetic data, however, the effects of these strategies are substantially weaker. Similar compression appears for affirmation and suggestions, which are followed by distress reduction less often in the controlled synthetic conditions than in ESConv , shown in Table~\ref{tab:strategy_summary}. 
\textcolor{revisionblue}{
These differences are statistically significant for
affirmation in nine of ten comparison corpora
(Holm-adjusted $p<0.05$;
Appendix~\ref{app:significance_strategy_effects}),
including all three controlled synthetic conditions
($p_{\mathrm{Holm}}<0.01$).
}

%\lareina{would add two other exampels here} \sara{no space. added reference instead}

%\lareina{moved to appendix. I think we might mention this is similar to what we found with CTL instead} %\ndnote{I would think you'd probably want to explicitly reference that this is similar to the other paper/distinguish why this qualitative analysis is different or adds to the previous one unless we want to remove it from the CTL paper.} %\ndnote{To clarify, I think in particular it might be odd that we are using the same synthetic example for both the  CTL data and this paper, especially if its the only qualitative example. Would either choose a different one that gets the point across for one of the two papers, or at least add additional examples.} \lareina{agree, will use another example}

\begin{table}[t]
\centering
\scriptsize
\setlength{\tabcolsep}{2.6pt}
\renewcommand{\arraystretch}{1.08}

\begin{tabular}{@{}l*{8}{rr}@{}}
\toprule
& \multicolumn{2}{c}{\textbf{Question}}
& \multicolumn{2}{c}{\textbf{Affirm.}}
& \multicolumn{2}{c}{\textbf{Reflect.}}
& \multicolumn{2}{c}{\textbf{Restate}}
& \multicolumn{2}{c}{\textbf{Suggest.}}
& \multicolumn{2}{c}{\textbf{Info.}}
& \multicolumn{2}{c}{\textbf{Self-disc.}}
& \multicolumn{2}{c}{\textbf{Others}} \\
\cmidrule(lr){2-3}
\cmidrule(lr){4-5}
\cmidrule(lr){6-7}
\cmidrule(lr){8-9}
\cmidrule(lr){10-11}
\cmidrule(lr){12-13}
\cmidrule(lr){14-15}
\cmidrule(lr){16-17}

\textbf{Corpus}
& \textbf{Use} & \textbf{$\downarrow$Dist.}
& \textbf{Use} & \textbf{$\downarrow$Dist.}
& \textbf{Use} & \textbf{$\downarrow$Dist.}
& \textbf{Use} & \textbf{$\downarrow$Dist.}
& \textbf{Use} & \textbf{$\downarrow$Dist.}
& \textbf{Use} & \textbf{$\downarrow$Dist.}
& \textbf{Use} & \textbf{$\downarrow$Dist.}
& \textbf{Use} & \textbf{$\downarrow$Dist.} \\
\midrule

\rowcolor{gray!15}
\multicolumn{17}{@{}l}{\textbf{\textit{Real}}} \\
ESConv
& 23.8 & 30.4
& 20.0 & 45.1
& 3.4 & 39.2
& 3.1 & 30.1
& 14.5 & 48.2
& 3.3 & 46.4
& 11.9 & 45.5
& 20.0 & 33.9 \\

\midrule
\rowcolor{gray!15}
\multicolumn{17}{@{}l}{\textbf{\textit{Semi-real}}} \\
AnnoMI
& 20.6 & 29.5
& 17.2 & 28.4
& 1.4 & 33.3
& 10.0 & 32.5
& 5.8 & 32.8
& 3.0 & 24.1
& 1.2 & 34.9
& 40.9 & 24.2 \\

PsyDial
& 25.0 & 34.3
& 27.5 & 41.2
& 14.8 & 42.3
& 10.2 & 37.4
& 8.6 & 41.4
& 2.1 & 39.1
& 1.2 & 39.0
& 10.5 & 33.6 \\

\midrule
\rowcolor{gray!15}
\multicolumn{17}{@{}l}{\textbf{\textit{Synthetic}}} \\
PsychEval
& 47.4 & 33.7
& 2.0 & 50.0
& 0.3 & 42.1
& 1.0 & 36.2
& 46.6 & 41.1
& 0.9 & 49.2
& 0.1 & 36.9
& 1.7 & 43.8 \\

CACTUS
& 26.2 & 31.9
& 25.2 & 37.2
& 1.6 & 33.2
& 10.4 & 33.1
& 32.5 & 43.8
& 0.6 & 35.7
& 0.0 & --
& 3.5 & 26.4 \\

SimPsyDial
& 17.6 & 37.5
& 17.1 & 41.8
& 2.1 & 41.5
& 7.2 & 38.6
& 49.0 & 46.5
& 0.8 & 36.8
& 0.1 & --
& 6.0 & 29.5 \\

\textcolor{revisionblue}{PsyChain}
& \textcolor{revisionblue}{7.4} & \textcolor{revisionblue}{38.2}
& \textcolor{revisionblue}{23.8} & \textcolor{revisionblue}{39.1}
& \textcolor{revisionblue}{4.1} & \textcolor{revisionblue}{36.3}
& \textcolor{revisionblue}{14.6} & \textcolor{revisionblue}{36.4}
& \textcolor{revisionblue}{47.4} & \textcolor{revisionblue}{42.3}
& \textcolor{revisionblue}{2.2} & \textcolor{revisionblue}{37.2}
& \textcolor{revisionblue}{0.1} & \textcolor{revisionblue}{--}
& \textcolor{revisionblue}{0.4} & \textcolor{revisionblue}{--} \\

Baseline DA
& 8.8 & 33.5
& 44.4 & 35.8
& 3.5 & 29.1
& 5.4 & 34.8
& 33.4 & 33.2
& 0.1 & --
& 0.0 & --
& 4.5 & 35.3 \\

Dual Agent
& 4.7 & 32.8
& 47.5 & 37.1
& 2.4 & 36.3
& 2.3 & 38.1
& 38.2 & 40.7
& 0.1 & --
& 0.0 & --
& 4.9 & 48.3 \\

Baseline ZS
& 17.3 & 28.8
& 43.8 & 38.2
& 2.6 & 41.3
& 8.6 & 32.7
& 17.8 & 42.7
& 2.8 & 38.3
& 2.5 & 16.2
& 4.6 & 40.7 \\

Zero-shot
& 14.1 & 34.3
& 50.1 & 39.4
& 2.3 & 40.2
& 5.2 & 34.7
& 17.9 & 42.4
& 2.6 & 37.4
& 1.8 & 35.1
& 6.1 & 29.4 \\

\bottomrule
\end{tabular}

\caption{
Counselor strategy and downstream distress change across datasets. 
% \textcolor{revisionblue}{PsyChain included.}
For each strategy, \textbf{Use} is the percentage of counselor turns assigned
to it; \textbf{$\downarrow$Dist.} is the percentage of uses
followed by lower patient distress. All values are percentages.
Missing values indicate insufficient observations.
}
\label{tab:strategy_summary}
\end{table}
% TODO (ND's note): if the companion CTL paper reports the same tailoring
% gap, add one sentence here saying this replicates it on public data and
% that the present analysis adds the protocol attribution; otherwise drop
% the case study from one of the two papers.

Besides strategy diversity, \textit{patient--counselor coupling} is also substantially weaker in synthetic data, as shown in Figure~\ref{fig:process_comparison}(c). In ESConv, both strategy and affect remain predictive of the patient's next state after controlling for conversation stage. However, in synthetic data, much of this dependence disappears, especially for SimPsyDial and Dual Agent, while no synthetic corpus approaches ESConv in emotion coupling. 
\textcolor{revisionblue}{
Stage-adjusted conditional mutual information confirms
these differences: all eight synthetic corpora exhibit
significantly weaker emotion coupling than ESConv
(Holm-adjusted $p<0.01$), while six of eight also
exhibit significantly weaker strategy coupling
(Appendix~\ref{app:significance_coupling}).
Notably, PsyChain shows reductions in
both channels despite its collaborative multi-agent
generation framework.
}
Apparent strategy coupling in CACTUS and PsychEval is better explained by plan-guided joint authorship, whereas two-agent systems often follow shared stage schedules rather than adapt turn by turn. Overall, synthetic strategy use and coupling primarily track prompting, staging, and authorship structure rather than genuine interpersonal adaptation.

% TODO: insert Baseline/Zero-shot stage-adjusted values; add PsychEval
% treatment stage as a permutation stratum.

\subsection{Does synthetic data capture the diversity of real recovery pathways?}
\label{sec:diversity}

\textbf{Synthetic datasets can exhibit trajectory-level mode collapse, and
more samples do not necessarily expand behavioral coverage.} From Figure~\ref{fig:process_comparison}(d) and the full archetypes distribution reported in
Appendix Table~\ref{tab:trajectory_patterns}, we see that
ESConv spans all five recovery archetypes and shows the greatest trajectory
diversity. 
\textcolor{revisionblue}{
All nine comparison corpora exhibit significantly
lower mean pairwise trajectory distances than ESConv
(Holm-adjusted $p<0.001$;
Appendix~\ref{app:significance_diversity}).
}
 In contrast, several synthetic datasets concentrate on a narrow
set of pathways: Baseline DA and Dual Agent are dominated by unresolved distress,
PsychEval by persistent moderate distress, and Zero-shot by steady
de-escalation. CACTUS and SimPsyDial achieve broader, more ESConv-like
archetype coverage, suggesting that structured plans, counseling stages, and
patient grounding can increase dynamic coverage. However, diversity across
representational layers can diverge: Baseline shows relatively high strategy-
and emotion-label diversity despite having the lowest trajectory diversity.
Thus, generating more conversations from a behaviorally narrow pipeline may
increase surface variation without expanding the underlying recovery pathways,
and broader archetype coverage still does not guarantee that those trajectories
emerge through faithful turn-by-turn dynamics.
%\sara{rethinking if there are some better RQs to put in the intro that's covering the areas of evaluation more holistically....}

%\msnote{I think this question and the next can be combined}

\subsection{Which Fidelity Gaps Can Protocol Design Fix?}
%\lareina{NOT FINISHED!! will update soon} 
\label{sec:scorecard}

\begin{table}[t]
\centering
\small
\setlength{\tabcolsep}{4pt}
\begin{tabular}{@{}llcccc@{}}
\toprule
\textbf{Measure} & \textbf{Metric}
& \textbf{Control}
& \textbf{Tuned}
& \textbf{ESConv}
& \textbf{Result} \\
\midrule

\rowcolor{gray!15}
\multicolumn{6}{@{}l}{\textbf{\textit{Tier 1: varies with protocol}}} \\

\multirow{4}{*}{\blackcirc{1} Recovery}
 & End positive (\%)            & 53.0 & 58.0 & 67.3 & \cmark$^{\dagger}$ \\
 & Recovery onset    & 0.26 & 0.38 & 0.56 & \cmark$^{\dagger}$ \\
 & Last-mile share              & 0.38 & 0.46 & 0.50 & \cmark \\
 & Post-onset persistence (\%) & 40.7 & 46.4 & 53.4 & \cmark$^{\dagger}$\\
\addlinespace[2pt]

\multirow{2}{*}{\blackcirc{2}\ Archetype coverage}
& JSD to ESConv            & 0.083 & 0.041 & 0.00 & \cmark\\
& Dominant archetype   & 42.0  & 36.0  & 30.0 & \cmark$^{\dagger}$ \\
\midrule

\rowcolor{gray!15}
\multicolumn{6}{@{}l}{\textbf{\textit{Tier 2: persists across prompt interventions}}} \\

\multirow{2}{*}{\blackcirc{3}\ Strategy mix}
& Strategy evenness ($J$)            & 0.62 & 0.64 & 0.89 & \xmark \\
& Top-2 strategy share (\%)          & 79.8 & 78.5 & 34.6 & \xmark \\

\blackcirc{4}\ Strategy-channel coupling
& Stage-adjusted CMI (bits)
& 0.007 & 0.013 & 0.039 & \xmark \\

\blackcirc{5}\ Emotion-channel coupling
& Stage-adjusted CMI (bits)
& 0.001 & 0.002 & 0.019 & \xmark \\

\multirow{2}{*}{\raisebox{0.8ex}{\blackcirc{6} Turn-level relief}}
 & Strategy-explained $\Delta d$ (\%) & 1.6    & 0.2    & 1.9    & \xmark \\
 & Mean $\Delta d$ per counselor turn  & $-0.009$ & $-0.009$ & $-0.041$ & \xmark \\

\bottomrule
\end{tabular}

\caption{
Summary of prompt-tuning experiments.
For each process-fidelity gap, we report the primary
evaluation metric under the control (original) prompt,
the corresponding tuned prompt, and the ESConv reference.
\
\textbf{Result} indicates whether prompt tuning showed
meaningful movement toward the real reference;
\cmark$^{\dagger}$ indicates moderate or partial
improvement given the limited sample size.
}
\label{tab:prompt_tuning_gpt4o}
\end{table}

%\msnote{Have these results been added yet? I think this is important to foreground in claims about protocol design because I'm not convinced by the analysis of existing datasets how much control over interactional responsiveness the design was trying to achieve.}
% \lareina{maybe add one sentence about why zeroshot not dual agent?}
% \msnote{I think you need to introduce the prompt-tuning setup first and motivate this choice of a direction a bit. For example, I would think that if you were trying to scaffold the process you might want to change the prompt based on the point in the conversation vs. just tuning one prompt up front} 
We first test the simplest way to address this gap: making the desired process more explicit in the generation prompt while holding the generation architecture fixed. This controlled ablation study lets us ask whether the missing dynamics primarily reflect underspecified instructions, or whether they remain difficult to recover even when directly targeted.
% To test whether the identified gaps can be reduced through simple prompt tuning, we conduct a controlled ablation study. 
For each targeted measure in Table~\ref{tab:prompt_tuning_gpt4o}, we generate
100 conversations with the original prompt and 100 with the corresponding
prompt-tuned variant, using the same GPT-4o~\citep{openai_gpt4o} model and
\textbf{zero-shot} generation. This setup allows us to vary global trajectory
instructions while holding the generation architecture fixed. The interventions
add targeted instructions for recovery dynamics, trajectory diversity,
counselor strategy balance, patient--counselor coupling, and turn-level relief,
while otherwise leaving the base generation setup unchanged. Specific prompt
modifications are provided in Appendix~\ref{app:tuned_prompts}.
%\sara{Use more straightforward titles. Also write a summary here to explain what we tested +reults.}

The answer is asymmetric. Our experiments reveal that \textbf{more detailed instructions substantially improve properties that can be specified globally}, when recovery occurs and how trajectories are distributed,  \textbf{but have little effect on interactional properties} that require one turn to respond contingently to the previous one.
% Our prompt-tuning experiments reveal that
% \textbf{global trajectory properties can be shifted toward human behavior through
% prompt-level control, whereas interactional realism remains largely
% unchanged.}
For \blackcirc{1} \textbf{recovery}, tuning moved several aspects of the
recovery arc toward ESConv: recovery onset shifted later (0.26$\rightarrow$0.38), last-mile share increased (0.38$\rightarrow$0.46),
post-onset persistence improved (40.7\%$\rightarrow$46.4\%),
and positive endings moved toward the human rate (53.0\%$\rightarrow$58.0\%). Some of these shifts are moderate and should be interpreted
cautiously given the limited ablation sample size. Likewise, \blackcirc{2}
\textbf{archetype coverage} became more human-like: Jensen--Shannon divergence
from the ESConv archetype distribution fell from 0.083 to 0.041, and the
dominant-archetype share decreased from 42\% to 36\%.
In contrast, interventions targeting local interaction produced little change.
For \blackcirc{3} \textbf{strategy mix}, strategy evenness remained nearly
unchanged (0.62$\rightarrow$0.64; ESConv 0.89), with affirmation and suggestions
still dominating. Figure~\ref{fig:p5_strategy_pies} shows the specific strategy distribution, which does not show obvious change after tuning.\blackcirc{4}--\blackcirc{5} \textbf{strategy- and
emotion-channel coupling} increased only slightly after stage adjustment and
remained far below ESConv, indicating that explicitly prompting contingent
responses did not recover human-like counselor--patient coordination.
Finally, \blackcirc{6} \textbf{turn-level relief} did not improve either. Strategy-explained distress change decreased from 1.6\% to 0.2\% (ESConv 1.9\%),
while mean downstream distress change was unchanged. 
Together, these results suggest that detailed prompting can better specify the global shape of a trajectory, but does not by itself reproduce the contingent turn-by-turn dynamics through which human support unfolds.

\textcolor{revisionblue}{
To assess cross-model generalizability, we extend
the prompt-tuning experiments to three additional
frontier models: GPT-6.1 Sol, Claude Sonnet 5.5,
and Gemini 3.1 Pro. Details provided in Appendix~E.7--E.9.
Across all three models, none of the
three models achieves a 15\% reduction in the gap
for strategy-channel coupling, emotion-channel
coupling, or turn-level relief. These results reinforce the finding that prompt-level
interventions can improve global recovery patterns
without comparably improving interactional dynamics.
}

% Together, these results
% suggest that prompting can reshape \emph{when} and \emph{how broadly} recovery
% occurs, but does not by itself reproduce the turn-by-turn dependence through
% which human support unfolds.
% \ndnote{Not that important, but I think there is some work that similarly discuss the problem of trying to prompt for diversity that might be useful to cite. Will see if I can find} % commenting out for submission!

% TODO: insert Table~\ref{tab:interventions} (scorecard measures under each
% intervention, with the unmodified condition as control) and replace the
% bracketed values below.

% If contingency does move emotion coupling, say so plainly and scope it:
% one prompt-level route to tier 2 exists in the two-agent setting; if it
% does not, state that tier 2 persisted under every intervention tested.

\paragraph{Synthetic Generation Ablations.}
% \lareina{not sure it we need to include this here, maybe briefly mention and put relevant figure to appendix?} \sara{keeping here}
In addition, we investigate additional ablation studies that help identify which artifacts persist across configurations
and what the dual-agent setup does, or does not, provide beyond single-pass
zero-shot generation. Generation failures involving length control,
termination, and role drift are documented in
Appendix~\ref{app:failure_modes}. Within the zero-shot corpus, we find that model choice has the largest effect. GPT-4o
conversations begin at lower distress, exhibit roughly twice the early-to-late distress reduction, and show greater volatility. By contrast, target length and topic specification have
little effect: short vs.\ long and specified vs.\ unspecified conversations
begin at nearly identical distress levels ($\approx 0.80$), and their
early-to-late changes differ by only about 0.03. Overall, the synthetic
emotional arc depends substantially more on the generator model than on
conversation length or topic specification.
% TODO: one paragraph of ablation results: which tier-1 measures move with
% model family / length / topic, and confirm tier-2 measures do not.

\section{Discussions and Conclusion}
\label{sec:discussion}

\paragraph{Discussion.}Our results suggest that process fidelity has two components: \emph{trajectory-level control} and \emph{interactional responsiveness}. Synthetic counseling generation reliably satisfies explicit trajectory constraints---when recovery occurs and which broad arc is followed---but is less able to specify the interactional dynamics within that arc. Detailed prompting can move several global properties toward the human reference, yet contingent turn-by-turn responsiveness remains difficult even when directly targeted. Across synthetic datasets, counselor behavior carries less information about the patient's subsequent state than in human support; moreover, apparent coupling may reflect shared schedules or joint authorship rather than genuine adaptation. Process-level measures should therefore complement endpoint metrics, and generation targets should be calibrated to real conversational distributions rather than idealized positive trajectories. Future work should explore trajectory and emotion labels as process-level supervision,
including reinforcement learning or trajectory-aware reward functions, and develop
generation methods that explicitly model intervention-conditioned patient state
transitions rather than relying on shared dialogue-level schedules.

\paragraph{Conclusion.}
We introduced a process-centered framework for evaluating whether synthetic counseling dialogues reproduce not only plausible outcomes, but also the trajectories through which those outcomes emerge. Across ten human, reconstructed, and synthetic datasets, we find that generation protocols can control many outcome- and timing-level properties, while turn-by-turn responsiveness remains consistently difficult to reproduce. Synthetic dialogue can therefore reach the right destination while following the wrong journey.

\subsection*{AI use statement}

Generative AI systems are the object of study in this work: the synthetic
dialogues we generate and evaluate, the translation of Chinese datasets, and
the emotion-labeling pipeline all use the models named in
\S\ref{sec:datasets}--\S\ref{sec:methods}, and these uses are documented
there. We also used generative AI tools for writing assistance. All AI-assisted
text was reviewed and revised by the authors, and all analyses, results, and
claims were independently checked by the authors. We take responsibility for
the final content of this work, including text, claims, analyses, and artifacts
produced with the aid of generative AI.
% TODO: fill the bracketed disclosures per the ICLR 2027 AI Policy.

\subsection*{Ethics statement}
% \ndnote{I probably wouldn't call out "crisis-line data" specifically unless all the datasets you use are crisis-line convos}
All datasets are public releases or generated data used under their respective licenses; no human subjects were recruited for this study. The
synthetic datasets contain no real individuals, and the semi-real datasets
were de-identified by their creators. We do not release new dialogue
content beyond what is needed to reproduce our analyses. Our findings
caution against training or evaluating deployed support systems on synthetic
dialogue whose process fidelity has not been checked, and we intend the
framework as a safeguard for that use. Because PsyDial and PsychEval are translated from Mandarin before emotion labeling, translation may smooth or alter affective distinctions or may not perfectly capture differences across cultures; cross-language comparisons should therefore be interpreted cautiously. 

\subsection*{Reproducibility statement}

Dataset provenance and preprocessing are described in
Appendix~\ref{app:dataset_details}; generation schemes, ablation conditions,
and prompts in Appendix~\ref{sec:generation_conditions}; and metric definitions in \S\ref{sec:methods}
and Appendix~\ref{app:volunteer_coupling}. Code, generated conversations, and per-utterance labels are released at the anonymous repository.\footnote{\url{https://anonymous.4open.science/r/synthetic-counseling-data-D356}}

\subsubsection*{Acknowledgments}
This work is in part supported by the funds provided by the National Science Foundation and by
DoD OUSD (R\&E) under Cooperative Agreement
PHY-2229929 (The NSF AI Institute for Artificial
and Natural Intelligence). The views, opinions
and/or findings expressed are those of the authors
and should not be interpreted as representing the
official views or policies of the National Science
Foundation or the U.S. Government.

\bibliography{iclr2027_conference}
\bibliographystyle{iclr2027_conference}

\appendix
\section{Dataset Details}
\label{app:dataset_details}

This appendix provides detailed descriptions of the datasets summarized
in Table~\ref{tab:datasets}, including their provenance,
interaction structure, and preprocessing for our analysis.

% ======================================================================
% Appendix A: Dataset Details (revised)
% Requires \usepackage{booktabs} for Table~\ref{tab:provenance}.
% New/changed bib keys are listed in appendix_datasets_refs.bib.
% ======================================================================

\subsection{Real Datasets}

\paragraph{ESConv.}
We include ESConv~\citep{liu-etal-2021-towards} as our public real-data anchor
for human--human emotional support. In ESConv, a patient discusses a
personal difficulty with a peer supporter in a live text conversation.
Supporters were crowdworkers trained using the Emotional Support
Conversation framework, a three-stage model organized around
exploration, comfort, and action.

Supporter utterances were originally annotated with eight strategies:
Question, Affirmation and Reassurance, Reflection of Feelings,
Restatement, Providing Suggestions, Information, Self-disclosure, and
Others. We adopt this inventory as our study-wide counselor-strategy
taxonomy. The public release contains 1{,}300 conversations. After
conversion to our unified dialogue schema, sessions contain an average
of 29.5 utterances.

ESConv is not a crisis-counseling corpus: its supporters are trained
peer crowdworkers rather than professional counselors or crisis
counselors, and its conversations are relatively short. We therefore
treat ESConv as a conservative public anchor for authentic support
interaction, rather than as a substitute for private crisis-line data.
The peer role also matters for interpretation. Behaviors such as
self-disclosure and small talk are common in peer support but are
discouraged by professional counseling norms, which every other
corpus in our comparison (real or simulated) adopts. Differences in
these behaviors therefore partly reflect supporter role rather than
data origin alone (\S\ref{sec:strategy}).

\subsection{Semi-real Datasets}

\paragraph{AnnoMI.}
AnnoMI~\citep{wu2023annomi} is a publicly available corpus of 133
professionally transcribed and expert-annotated motivational
interviewing (MI) sessions. The sessions are \emph{performed
demonstrations} drawn from educational and training videos rather than
naturally occurring clinical encounters: 110 demonstrate high-quality
MI and 23 demonstrate low-quality MI. Because the dialogues are
human-produced but staged, and pass through transcription,
segmentation, and annotation, we group AnnoMI with the semi-real
datasets.

Two properties of AnnoMI shape its affective profile. First, MI targets
motivation for behavior change (e.g., substance use), so many sessions
do not open from acute emotional distress. Second, the transcripts
preserve spoken-language features such as disfluencies and short
backchannels (e.g., ``Right.'', ``Okay.''), which yield many
affectively neutral patient turns. After conversion to our turn-level
schema, conversations contain an average of 72.9 utterances; 130
conversations retain at least two affect-bearing patient turns and enter
the trajectory analyses.
% TODO: confirm that the 133 -> 130 drop is due to the >=2 labeled
% patient-turn criterion in Sec. 4.1.
% TODO (optional): if you run the high/low-quality split, add:
% "We retain the session-level quality label and report stratified
% results in Appendix X."

\paragraph{PsyDial.}
PsyDial~\citep{qiu-lan-2025-psydial} is a Chinese counseling corpus of
2{,}382 long-form dialogues reconstructed from Xinling, a private corpus
of real text-based counselor--patient sessions collected on a Chinese
online counseling platform~\citep{li2023understanding}. It is built
with the RMRR pipeline (Retrieve, Mask, Reconstruct, Refine), which
treats the two speakers asymmetrically:
\begin{itemize}
    \item \textbf{Patient side (fully regenerated).} Every patient
    utterance in the source dialogue is masked. For each dialogue, the
    three most similar chief complaints are retrieved by dense retrieval
    from public PsyQA help-seeking posts~\citep{sun-etal-2021-psyqa} longer than
    300 Chinese characters. GPT-4o then rewrites all patient turns as a
    cloze task, conditioned on a retrieved complaint and on the full
    masked dialogue.
    \item \textbf{Counselor side (human-originated, edited).} Counselor
    utterances come from the real sessions. Expert counselors reviewed
    them to replace identifying details, and an LLM then refined them
    for fluency with respect to the reconstructed patient turns, subject
    to a similarity filter that keeps patient turns fixed.
\end{itemize}
The released PsyDial corpus corresponds to the refined version (D4 in
the original paper).

PsyDial is therefore best described as \emph{real counselor, synthetic
patient}. This has three implications for our analyses. (i) All patient
distress trajectories in PsyDial are produced by an LLM whose persona
comes from a different individual's post, not by the original patient.
(ii) counselor turns preserve the timing and strategy choices of
professional counselors, although their wording was lengthened during
refinement. (iii) Because each patient turn is written with the entire
masked dialogue visible, including later counselor turns, and counselor
turns are subsequently refined to fit the new patient turns, part of
PsyDial's patient--counselor dependence arises by construction. PsyDial
remains an informative middle case: it tests whether anchoring an LLM
patient to real counselor behavior recovers dynamics that are lost in
fully simulated dialogue. In our processed version, conversations
contain an average of 75.6 utterances.

\subsection{External Synthetic Datasets}

\paragraph{CACTUS.}
We include CACTUS~\citep{lee-etal-2024-cactus} as an external synthetic
corpus. The complete dataset contains 31{,}577 fully LLM-generated,
single-session counselor--patient dialogues built in three steps.
\begin{itemize}
    \item \textbf{Patient conditioning.} Personas, negative thoughts,
    and cognitive-distortion patterns are taken from
    PatternReframe~\citep{maddela-etal-2023-training}. GPT-3.5 converts each
    into a first-person intake form. Each patient is also assigned one of
    three attitudes (positive, neutral, negative), each with a list of
    associated behaviors.
    \item \textbf{Counselor planning.} Given the negative thought, its
    patterns, and the \emph{target reframed thought} from
    PatternReframe, GPT-3.5 selects three CBT techniques and writes a
    counseling plan directed at that reframing.
    \item \textbf{Dialogue generation.} GPT-4o writes the complete
    dialogue in a single pass (``script mode''), authoring both roles.
    The authors chose this over two-agent generation after human
    evaluation found scripted dialogues more natural. The prompt
    instructs the counselor not to name techniques or impose positive
    affirmations.
\end{itemize}
Generated dialogues were then filtered to 20--35 turns and to an average
score of at least 5 (on a 0--6 scale) across six Cognitive Therapy
Rating Scale (CTRS) items judged by GPT-3.5, retaining 86.3\% of
candidates. The corpus targets mild depression and anxiety. Among our
comparison sets, CACTUS is structurally closest to our Baseline and
Zero-shot conditions, since a single model authors both speakers.

The raw release contains 34 distinct \texttt{cbt\_technique} strings,
including typographical variants, formatting variants, synonyms, and
one multi-technique label. We normalize these strings into 18 canonical
techniques.
% TODO: CACTUS defines 12 techniques in its generation prompt. Check
% whether some of the 18 canonical labels can be merged further.
We then draw a stratified sample over the attitude~$\times$~technique
grid using proportional allocation at the corpus-level sampling rate,
with a minimum of 10 conversations for each non-empty cell. This
procedure yields 2{,}173 conversations and 66{,}368 utterances,
corresponding to an average of 30.5 utterances per conversation.

\paragraph{PsychEval.}
PsychEval~\citep{pan-etal-2026-psycheval} is a multi-session Chinese
counseling benchmark synthesized from 369 clinical case reports
published in peer-reviewed Chinese psychology journals. The reports
span five therapeutic modalities plus an integrative category:
cognitive-behavioral (148), psychodynamic (50),
humanistic-existential (50), postmodernist (50), behavioral (43), and
integrative (28). Only 40\% of cases are therefore CBT-based.

Generation proceeds in two phases, using GPT-5 with high reasoning
effort.
\begin{itemize}
    \item \textbf{Structured case extraction.} Each report is converted
    into a patient profile, a three-stage multi-session treatment plan,
    verbatim dialogue excerpts used as style exemplars, and links to a
    hierarchical skill library (677 meta-skills and 4{,}577 atomic
    skills).
    \item \textbf{Session-by-session generation.} Each session proceeds
    through three steps. Before the session, candidate skills are
    retrieved and session objectives are set. During the session, a
    single model writes the complete dialogue, playing both counselor
    and patient, and produces a reasoning trace (patient assessment,
    patient state, selected skill, strategy) before every counselor
    turn. After the session, the pipeline writes a summary, updates the
    profile of information the counselor has learned, and plans the
    next session.
\end{itemize}
Several prompt constraints are directly relevant to our metrics.
\begin{itemize}
    \item \textbf{Patient behavior.} The patient is instructed to display
    non-linear emotions, at least one form of mild resistance on
    difficult topics, and one or two topic drifts per session.
    \item \textbf{Counselor behavior.} The counselor is limited to at
    most one question per utterance, is told to use Socratic
    questioning frequently, and must close each session with a summary
    and homework.
    \item \textbf{Length.} Sessions target 25--70 exchanges.
\end{itemize}
Psychology researchers audited samples of the output, and deficient
segments were regenerated or manually corrected.

Each case contains 6--10 sessions progressing through three stages
(case conceptualization, core intervention, and consolidation), and
the pipeline is designed for improvement \emph{across} sessions rather
than within a single session. Because our unit of analysis is a single
conversation, we treat each session as an independent conversation,
yielding 2{,}798 session-level samples with an average of 49.1
utterances per session. We remove the reasoning traces so that our
analysis includes only the spoken dialogue.
% TODO: if you keep session index / stage, add: "We retain each
% session's treatment stage for stratified analyses (Appendix X)."

\paragraph{SimPsyDial.}
We also include SimPsyDial~\citep{qiu-lan-2026-interactive}, a synthetic
Chinese counseling corpus generated through LLM-to-LLM role play. Both
the patient and counselor agents are zero-shot prompted
\texttt{gpt-4-1106-preview} models.
\begin{itemize}
    \item \textbf{Patient agent.} Each of the 1{,}000 patients is
    conditioned on a distinct PsyQA help-seeking post longer than 300
    Chinese characters~\citep{sun-etal-2021-psyqa}. The patient is told to
    speak in one or two sentences, to disclose its problems step by
    step, and not to thank the counselor or say goodbye early. The
    prompt specifies no resistance behaviors.
    \item \textbf{Counselor agent.} The counselor uses the fixed persona
    ``Xiao Tian'' and follows an integrative three-stage model
    (exploration, insight, action) combining patient-centered,
    psychodynamic, and cognitive-behavioral techniques. It is instructed
    to ask one question at a time and to avoid empathy, restatement,
    and affirmation in the early stage. In the late stage, it asks the
    patient to summarize improvements.
    \item \textbf{Dialogue control.} The patient always opens with
    ``Hello.'' A conversation ends after at most 50 turns, or earlier
    when a counselor utterance matches a predefined list of farewell
    keywords. Counselor responses longer than 200 characters or
    containing line breaks or enumerations are rejected.
\end{itemize}
The public release contains 1{,}000 single-session conversations, with
an average of 25.9 utterances in our processed schema. Although its
patient profiles are grounded in authentic self-reported distress
narratives, all conversational turns are machine-generated and no human
participates during dialogue generation.

SimPsyDial is architecturally similar to our Dual Agent condition
because both employ independently prompted patient and counselor agents
that interact without turn-level human supervision. They differ in
profile source, underlying model, prompt design, and termination
mechanism.

\subsection{Generation Methods Summary}

Table~\ref{tab:provenance} summarizes who authors each side of the
dialogue in every corpus. This distinction matters for the coupling
analysis (\S\ref{sec:strategy}). When a single model writes both roles,
or when patient turns are written with later counselor turns visible,
counselor--patient dependence can arise from authorship rather than from
interaction.

\begin{table}[h]
\centering
\small
\begin{tabular}{llll}
\toprule
Corpus & counselor turns & patient turns & Joint authorship \\
\midrule
ESConv & Human (peer) & Human & No \\
AnnoMI & Human (performed) & Human (performed) & Scripted demo \\
PsyDial & Human, LLM-refined & LLM (GPT-4o) & Partial (cloze) \\
\midrule
CACTUS & LLM (GPT-4o) & LLM (GPT-4o) & Yes (script) \\
PsychEval & LLM (GPT-5) & LLM (GPT-5) & Yes (script) \\
SimPsyDial & LLM (GPT-4) & LLM (GPT-4) & No (two agents) \\
Baseline (Ours) & LLM & LLM & Yes (single model) \\
Dual Agent (Ours) & LLM & LLM & No (two agents) \\
Zero-shot (Ours) & LLM & LLM & Yes (single model) \\
\bottomrule
\end{tabular}
\caption{Authorship of counselor and patient turns in each corpus.
``Joint authorship'' indicates whether the patient's next turn is
written with knowledge of, or by the same process as, the counselor
turn it follows.}
\label{tab:provenance}
\end{table}

\section{Synthetic Data Generation Conditions}
\label{sec:generation_conditions}
\label{app:generation}

We compare three generation models:
GPT-4o (\texttt{gpt-4o-2024-08-06})~\citep{openai_gpt4o},
Claude Haiku 4.5 (\texttt{claude-haiku-4-5-20251001})~\citep{anthropic_haiku45},
and Gemini 2.5 Pro (\texttt{gemini-2.5-pro})~\citep{google_gemini25pro}.

\paragraph{Dual-Agent Setup. }
Two LLM agents, one conditioned as the patient and the other as the counselor, converse in alternating turns. The patient agent is prompted as a person in crisis with help-seeker constraints (reluctance to open up, skepticism toward unsolicited advice, disorganized affect, SMS-style register), and the counselor agent is prompted as a counselor following active-listening principles (validation, deepening questions, non-directive pacing). Each conversation opens with a fixed system preamble and a scripted handoff, then alternates until a closing-signal detector fires on farewell or wind-down cues; a maximum cap of 150 turns prevents runaway generation.

\paragraph{Zero-shot Setup. }
In the zero-shot setup, we prompt one LLM to generate an entire conversation in a structured output. We adapt the role-specific prompts used in the dual-agent setup with additional conversation length and flow-control instructions to specify the conversation schema, structural segments (preamble, exploration, closing), and a target length range. The model produces all messages, timestamps, and metadata in a single call, with a retry mechanism for failed generations. Unlike the dual-agent setup, the zero-shot generator has no closing-signal detector or turn cap---conversation length is governed entirely by the prompt and the model's own termination.
% TODO: include the patient prompt, the counselor prompt (first principle:
% validation), and the Baseline prompt verbatim; the main text cites
% Appendix~\ref{app:generation} for all three.

\begin{table}[h]
\centering
\footnotesize
\setlength{\tabcolsep}{4pt}
\begin{tabular}{llllll}
\toprule
\textbf{ID} & \textbf{Method} & \textbf{Models} & \textbf{Topic} & \textbf{Ending} & \textbf{Ending Way} \\
\midrule
G1-A & Dual-agent & gpt4o / gpt4o   & Unspecified & Natural & No suggestion \\
G1-B & Dual-agent & Claude / Claude & Unspecified & Natural & No suggestion \\
G1-C & Dual-agent & Gemini / Gemini & Unspecified & Natural & No suggestion \\
G1-D & Dual-agent & Claude / gpt4o  & Unspecified & Natural & No suggestion \\
G1-E & Dual-agent & Claude / Gemini & Unspecified & Natural & No suggestion \\
G1-F & Dual-agent & Gemini / gpt4o  & Unspecified & Natural & No suggestion \\
G1-G & Zero-shot  & gpt4o           & Unspecified & Natural & No suggestion \\
G1-H & Zero-shot  & Claude          & Unspecified & Natural & No suggestion \\
G1-I & Zero-shot  & Gemini          & Unspecified & Natural & No suggestion \\
\bottomrule
\end{tabular}
\caption{Group G1 generation settings for the initial method and model comparison.}
\label{tab:g1_conditions}
\end{table}

\begin{table}[h]
\centering
\footnotesize
\setlength{\tabcolsep}{4pt}
\begin{tabular}{llllll}
\toprule
\textbf{ID} & \textbf{Models} & \textbf{Topic} & \textbf{Ending} & \textbf{Ending Way} & \textbf{Turns} \\
\midrule
G2-A1 & Claude / gpt4o & Unspecified        & Natural          & No suggestion         & Unknown \\
G2-A2 & Claude / gpt4o & Unspecified        & Natural          & Within prompt (range) & 60--80 \\
G2-A3 & Claude / gpt4o & Specific disorders & Natural          & No suggestion         & Unknown \\
G2-A4 & Claude / gpt4o & Specific disorders & Natural          & Within prompt (range) & 60--80 \\
G2-A5 & Claude / gpt4o & Unspecified        & Enforced closing & Enforce               & 60--80 \\
G2-A6 & Claude / gpt4o & Unspecified        & Enforced closing & Enforce               & 120--150 \\
G2-A7 & Claude / gpt4o & Specific disorders & Enforced closing & Enforce               & 60--80 \\
G2-A8 & Claude / gpt4o & Specific disorders & Enforced closing & Enforce               & 120--150 \\
\bottomrule
\end{tabular}
\caption{Group G2 controlled ablations using the Claude/gpt4o dual-agent setting.}
\label{tab:g2_conditions}
\end{table}

\begin{table}[h]
    \centering
    \footnotesize
    \setlength{\tabcolsep}{4pt}
    \begin{tabular}{llllll}
    \toprule
    \textbf{ID} & \textbf{Model} & \textbf{Topic} & \textbf{Ending} & \textbf{Ending Way} & \textbf{Turns} \\
    \midrule
    G3-B1  & gpt4o      & Unspecified        & Natural & Within prompt (range) & 60--80 \\
    G3-B2  & Claude 4.5 & Unspecified        & Natural & Within prompt (range) & 60--80 \\
    G3-B3  & Gemini     & Unspecified        & Natural & Within prompt (range) & 60--80 \\
    G3-B4  & gpt4o      & Specific disorders & Natural & Within prompt (range) & 60--80 \\
    G3-B5  & Claude 4.5 & Specific disorders & Natural & Within prompt (range) & 60--80 \\
    G3-B6  & Gemini     & Specific disorders & Natural & Within prompt (range) & 60--80 \\
    G3-B7  & gpt4o      & Unspecified        & Natural & Within prompt (range) & 120--150 \\
    G3-B8  & Claude 4.5 & Unspecified        & Natural & Within prompt (range) & 120--150 \\
    G3-B9  & Gemini     & Unspecified        & Natural & Within prompt (range) & 120--150 \\
    G3-B10 & gpt4o      & Specific disorders & Natural & Within prompt (range) & 120--150 \\
    G3-B11 & Claude 4.5 & Specific disorders & Natural & Within prompt (range) & 120--150 \\
    G3-B12 & Gemini     & Specific disorders & Natural & Within prompt (range) & 120--150 \\
    \bottomrule
    \end{tabular}
    \caption{Group G3 zero-shot ablations over model, topic control, and target turn range.}
    \label{tab:g3_conditions}
\end{table}

Tables~\ref{tab:g1_conditions}--\ref{tab:g3_conditions} summarize the synthetic data generation settings used in our ablation study.

\subsection{Generation Failure Modes}
\label{app:failure_modes}

Generating high-quality synthetic conversations required addressing
several non-trivial generation failures:

\begin{itemize}
    \item \textbf{Zero-shot conversation-length control.}
    Without structural guidance, the zero-shot generator frequently
    violated the requested conversation length by either terminating
    prematurely or producing excessively long exchanges. Adding an
    explicit dialogue structure to the prompt
    substantially improved adherence to the target length.
    % TODO: restore \ref{app:zeroshot_prompt_unspecified} once the prompt
    % is included in this appendix.

    \item \textbf{Dual-agent termination failure.}
    The interacting agents frequently failed to identify natural
    stopping points, despite receiving the complete conversation history
    and explicit length instructions. Consequently, conversations often
    continued until they were externally stopped at the maximum turn
    limit, indicating that conversational closure was not reliably
    inferred from context alone.

    \item \textbf{Dual-agent role confusion.}
    The patient agent occasionally drifted into the counselor role and
    produced responses inconsistent with its assigned persona. This
    behavior typically emerged after several turns, suggesting that role
    grounding weakened as the interaction history grew.
\end{itemize}

\subsection{Synthetic Dialogue Prompts}
\label{app:synthetic_prompts}

\paragraph{Zero-shot prompt.}
\label{app:zeroshot_prompt_unspecified}
The single-pass generator receives the template below; provider and
model name are filled per run.

\begin{promptbox}{Zero-Shot Generation}
\textbf{Prompt:} Generate a realistic Crisis Text Line conversation about a
realistic emotional or psychological challenge that someone might reach
out to a crisis line about. Choose ONE specific topic and focus the
conversation on that primary concern while allowing natural tangents and
related issues to emerge.

CRITICAL: This conversation must be LONG. Count the messages as you
generate. Do NOT conclude before message 60. The middle exploration phase
should be extensive---at least 50 exchanges before any wrap-up begins.
Take time to explore different facets of the patient's situation before
moving toward closure.

STRUCTURE:
\begin{itemize}[nosep,leftmargin=*]
\item Opening (messages 1--6): System preamble + patient intro
\item Exploration (messages 7--60): Extended back-and-forth. DO NOT RUSH.
\item Closing (messages 61--76): Gradual wrap-up
\item System tail (messages 77--80): Fixed closing sequence
\end{itemize}

REMINDER AT MESSAGE 30: You are only halfway. Continue exploring new
angles. REMINDER AT MESSAGE 50: Begin thinking about gradual closure, but
not yet.

OUTPUT ONLY THE JSON OBJECT---no markdown, no code fences, no commentary.
The conversation must follow this EXACT JSON schema:

{\ttfamily\small
\{\ "conversation\_id": "C\_2025\_XXXXXX", "topic\_seed": "mental\_health",
"created\_at": "YYYY-MM-DDTHH:MM:SSZ",
"model": \{"provider": "\textless PROVIDER\textgreater", "name":
"\textless MODEL\_NAME\textgreater", "temperature": 0.7\},
"meta": \{"natural\_flow": true, "min\_gap\_minutes": 1,
"max\_gap\_minutes": 5\},
"messages": [\{"author": "System", "actor\_id": "1", "datetime":
"MM/DD/YYYY HH:MM", "message\_id": 1, "text": "message text",
"notes": ""\}],
"resources": [\{"type": "url", "value": "\{\{URL\}\}"\},
\{"type": "email", "value": "\{\{EMAIL\}\}"\}]\ \}
}

CONVERSATION STRUCTURE (MUST FOLLOW):
\begin{enumerate}[nosep,leftmargin=*]
\item System preamble (2 messages): ``Thanks for texting Crisis Text
Line, where you'll text with a compassionate Crisis Counselor. Our terms
of service: crisistext.org/privacy'' and ``Text STOP anytime to cancel.
We can't receive msgs longer than 160 characters ($\sim$ a couple of
sentences). What's on your mind?''
\item patient opens with distress related to the ONE chosen topic
($\le$160 chars).
\item System: ``Thanks for sharing. We're getting a Crisis Counselor for
you, it may take a moment.''
\item counselor introduction: ``Hello, my name is [scrubbed] and I'm here
to listen.'' ``I'm a trained Crisis Counselor, but not a medical
professional.'' One gentle opening question.
\item Middle section (extended natural flow): Alternate between patient
and counselor, with SHORT patient messages ($\le$160 chars) and counselor
replies of 1--3 sentences. Keep all responses short and to the point,
using simple, human language rather than formal, clinical, or overly
polished phrasing. The middle should contain the vast majority of the
messages, with multiple waves of sharing, clarifying, circling back to
earlier themes, and gradually deepening understanding across many angles
of the patient's situation. For counselor, do not repeat the patient's
exact words; show understanding using fresh language. For patient, express
reluctance to open up at first; avoid showing high insight or prior
therapy experience. For patient, when describing distress, express
emotions and thoughts in a disorganized and emotional manner. When
presented with suggestions, the patient may show skepticism or uncertainty
and question feasibility. For counselor, stay gentle and nonjudgmental;
do not push the patient to open up faster than they choose. Occasionally
allow the patient to send 2 messages back-to-back. Let the conversation
end naturally when the patient signals they are ready to stop.
\item System closing tail (ALWAYS appended after the final human
exchange): System: ``Thanks for texting Crisis Text Line. The Crisis
Counselor has closed the conversation. You got this!'' System: ``Hey---We're
always looking to do the best we can. Did you find this conversation
helpful? Y/N'' patient: ``Y'' or ``N'' System: ``Thanks! Answer a few
more questions to help us understand how to better help others like you:
\{\{URL\}\}''
\end{enumerate}

CRITICAL REQUIREMENTS: patient messages MUST be $\le$160 characters.
Message IDs strictly increase. Timestamps progress by 1--5 minutes each
message. Use placeholders \{\{URL\}\}, \{\{EMAIL\}\} for resources and
survey link. The ``messages'' array MUST contain between 60 and 80
message objects total.
\end{promptbox}

\paragraph{Dual-agent prompts.}
\label{app:dual_agent_prompts}
The patient and counselor agents receive the templates below at every
turn, with the scenario, turn number, and conversation history filled in.

\begin{promptbox}{Dual-Agent patient}
\textbf{Prompt:} You are a person in crisis texting for support. Keep
responses natural and brief like real texting.

Scenario: \{\{chief\_complaint\}\} \quad Turn number: \{\{turn\_number\}\}
\quad Conversation so far: \{\{history\}\}

Instructions (FOLLOW EXACTLY):
\begin{enumerate}[nosep,leftmargin=*]
\item Keep your responses short and to the point.
\item Limit replies to 1--3 sentences.
\item Feel free to make up believable details from your past to answer
questions.
\item When discussing emotional difficulties, keep responses centered on
core feelings.
\item Express reluctance to open up at first; avoid showing high insight
or prior therapy experience.
\item When presented with suggestions, show skepticism or uncertainty;
question feasibility.
\item When expressing doubts or fears, avoid jumping to solutions;
articulate concerns and stay with the emotion.
\item Avoid being highly self-aware about your own patterns or problems.
\item When describing distress, express emotions and thoughts in a
disorganized and emotional manner.
\item Vary sentence length to create a more natural rhythm in dialogue.
Do not use ellipses. Write using complete thoughts; this is not the same
as complete sentences, but may be in a complete sentence. Do not use
overdramatic language or figurative language (no similes,
personification, or fancy words). Use text abbreviations, such as tbh,
lol, or ttyl, where it is appropriate and makes a conversation more
organic.
\item Do not repeat sentences or the same emotion words. When expressing
your feelings, sometimes be specific about your situation, sometimes
openly share your insecurities, and question your self-worth to convey a
deeper sense of vulnerability. Other times, push back and say that you
are not comfortable talking about something yet or feel embarrassed
talking about it; things that someone may not feel comfortable sharing
include sexual history, bad past experiences with family members, etc.
\item Sometimes ask for advice or help on how to solve your personal
problem, such as ``what can I do?''
\end{enumerate}

Return only the message text without any role labels or prefixes.
\end{promptbox}

\begin{promptbox}{Dual-Agent counselor}
\textbf{Prompt:} You are a crisis support counselor. You are responding
to a patient who needs help. Keep responses brief and natural.

Turn number: \{\{turn\_number\}\}
\begin{enumerate}[nosep,leftmargin=*]
\item Keep your responses short and to the point. Limit replies to 1--3
sentences.
\item Use simple, human language; avoid formal, clinical, or overly
polished phrasing.
\item Do not repeat the patient's exact words; show understanding using
fresh language.
\item Focus on their core feelings and show empathy about their
situations and emotions.
\item Avoid giving advice, reassurance, or solutions early in the
conversation, unless there is a clear safety concern.
\item Stay gentle and nonjudgmental; do not push the patient to open up
faster than they choose.
\item When exploring their experience, ask small, natural questions that
help them go one step deeper without pressure.
\item Vary your sentence structure and tone to feel organic; avoid canned
openings like ``It sounds like\ldots'' or ``I'm curious\ldots''
repeatedly.
\item Don't stay stuck on the same topic for more than 2--3 iterations;
guide the conversation forward when needed.
\item Keep the emotional focus on the patient; do not talk about yourself
or offer personal stories.
\item When reflecting feelings, keep statements brief, grounded, and
non-dramatic; avoid exaggerated empathy or big emotional language.
\item Maintain a steady, calm presence; validate their feelings in a
natural way.
\end{enumerate}

Conversation: \{\{history\}\}

Task: Write ONE brief counselor message that: (1) responds as the support
provider to someone in crisis; (2) responds naturally without repeating
back what they just said; (3) shows understanding through empathy, not
parroting their words; (4) asks follow-up questions that move the
conversation forward; (5) avoids advice/solutions early in the
conversation; (6) varies response style---sometimes validate, sometimes
explore, sometimes clarify; (7) lets the conversation flow like a real
text exchange.

Return only the message text without any role labels or prefixes.
\end{promptbox}

\section{Counselor Strategy Use and Coupling}
\label{app:volunteer_coupling}

We analyze three-turn windows $(u_t,h_t,u_{t+1})$, consisting of a
patient turn, the counselor response, and the subsequent patient turn.
We define the current and next patient states as
$X_t=e(u_t)$ and $Y_t=e(u_{t+1})$, the counselor strategy as
$S_t=s(h_t)$, and the counselor emotion as $A_t=e(h_t)$.

\paragraph{Immediate distress change.}
The downstream change in patient distress is
\[
\Delta d_t=d(u_{t+1})-d(u_t),
\]
where negative values indicate improvement. For each strategy, we
report its frequency, mean $\Delta d_t$ with conversation-level
bootstrap confidence intervals, and de-escalation rate
$\Pr(\Delta d_t<-0.1)$.

\paragraph{Conditional coupling.}
We measure whether the counselor turn predicts the patient's subsequent
state using conditional mutual information:
\[
\widehat I_{\mathrm{obs}}(Z;Y\mid X)
=
\sum_{x,z,y}\widehat p(x,z,y)
\log_2
\frac{\widehat p(y\mid x,z)}
     {\widehat p(y\mid x)},
\]
where $Z=S$ represents the strategy channel and $Z=A$ represents the
emotion channel. This measures the information supplied by the
counselor label beyond the patient's current state.

\paragraph{Permutation correction.}
To correct for finite-sample bias, we shuffle $Z_t$ among windows with
the same current patient state. Let $\overline I_{\mathrm{base}}$ denote
the mean CMI across these permutations. We define total coupling as
\[
\widehat I_{\mathrm{total}}
=
\widehat I_{\mathrm{obs}}
-
\overline I_{\mathrm{base}}.
\]

To remove dependence attributable to shared conversation timing, we
perform a stricter shuffle among windows with both the same current
patient state and the same conversation-progress quintile. Let
$\overline I_{\mathrm{stage}}$ denote this null mean. Stage-adjusted
coupling is
\[
\widehat I_{\mathrm{adj}}
=
\widehat I_{\mathrm{obs}}
-
\overline I_{\mathrm{stage}}.
\]

The difference between total and stage-adjusted coupling captures
dependence associated with both speakers following the same
conversation schedule. All coupling estimates represent predictive
dependence, not causal effects.

\section{Experiment Results}
\subsection{Emotion Transitions}

\begin{table}[H]
\centering
\caption{
Top-5 most frequent patient emotion transitions in each corpus.
Percentages are calculated over all transitions within each corpus.
}
\label{tab:top_emotion_transitions}
\scriptsize
\setlength{\tabcolsep}{3.5pt}
\renewcommand{\arraystretch}{1.08}

\resizebox{\textwidth}{!}{%
\begin{tabular}{
@{} l c l r r
@{\hspace{0.7cm}} l c l r r
@{\hspace{0.7cm}} l c l r r @{}
}
\toprule
\multicolumn{5}{c}{\textbf{ESConv (Real)}} &
\multicolumn{5}{c}{\textbf{AnnoMI (Semi-real)}} &
\multicolumn{5}{c}{\textbf{PsyDial (Semi-real)}} \\
\cmidrule(lr){1-5}
\cmidrule(lr){6-10}
\cmidrule(lr){11-15}
\multicolumn{3}{c}{\textbf{Transition}}
& \textbf{Count} & \textbf{\%} &
\multicolumn{3}{c}{\textbf{Transition}}
& \textbf{Count} & \textbf{\%} &
\multicolumn{3}{c}{\textbf{Transition}}
& \textbf{Count} & \textbf{\%} \\
\midrule

neutral & $\rightarrow$ & neutral & 836 & 4.70 &
neutral & $\rightarrow$ & neutral & 987 & 22.47 &
self-doubt & $\rightarrow$ & self-doubt & 2,242 & 2.62 \\

gratitude & $\rightarrow$ & gratitude & 556 & 3.13 &
self-doubt & $\rightarrow$ & self-doubt & 146 & 3.32 &
hopeful & $\rightarrow$ & hopeful & 2,238 & 2.61 \\

stress & $\rightarrow$ & stress & 488 & 2.75 &
self-aware & $\rightarrow$ & self-aware & 144 & 3.28 &
neutral & $\rightarrow$ & neutral & 2,213 & 2.59 \\

sadness & $\rightarrow$ & sadness & 434 & 2.44 &
self-aware & $\rightarrow$ & neutral & 110 & 2.50 &
self-aware & $\rightarrow$ & self-aware & 2,048 & 2.39 \\

disappointment & $\rightarrow$ & disappointment & 353 & 1.99 &
neutral & $\rightarrow$ & self-aware & 102 & 2.32 &
disappointment & $\rightarrow$ & disappointment & 1,401 & 1.64 \\

\midrule
\multicolumn{5}{c}{\textbf{PsychEval (Synthetic)}} &
\multicolumn{5}{c}{\textbf{CACTUS (Synthetic)}} &
\multicolumn{5}{c}{\textbf{SimPsyDial (Synthetic)}} \\
\cmidrule(lr){1-5}
\cmidrule(lr){6-10}
\cmidrule(lr){11-15}
\multicolumn{3}{c}{\textbf{Transition}}
& \textbf{Count} & \textbf{\%} &
\multicolumn{3}{c}{\textbf{Transition}}
& \textbf{Count} & \textbf{\%} &
\multicolumn{3}{c}{\textbf{Transition}}
& \textbf{Count} & \textbf{\%} \\
\midrule

neutral & $\rightarrow$ & neutral & 6,109 & 9.61 &
self-doubt & $\rightarrow$ & self-doubt & 3,590 & 11.82 &
self-doubt & $\rightarrow$ & self-doubt & 336 & 3.07 \\

self-aware & $\rightarrow$ & self-aware & 2,365 & 3.72 &
hopeful & $\rightarrow$ & hopeful & 1,592 & 5.24 &
hopeful & $\rightarrow$ & hopeful & 336 & 3.07 \\

anxiety & $\rightarrow$ & anxiety & 2,234 & 3.51 &
worthlessness & $\rightarrow$ & self-doubt & 769 & 2.53 &
self-aware & $\rightarrow$ & self-aware & 263 & 2.40 \\

self-aware & $\rightarrow$ & neutral & 1,674 & 2.63 &
self-doubt & $\rightarrow$ & hopeful & 765 & 2.52 &
fear & $\rightarrow$ & fear & 210 & 1.92 \\

neutral & $\rightarrow$ & self-aware & 1,608 & 2.53 &
anxiety & $\rightarrow$ & anxiety & 738 & 2.43 &
self-doubt & $\rightarrow$ & hopeful & 207 & 1.89 \\

\midrule
\multicolumn{5}{c}{\textbf{Baseline (Synthetic)}} &
\multicolumn{5}{c}{\textbf{Dual Agent (Synthetic)}} &
\multicolumn{5}{c}{\textbf{Zero-shot (Synthetic)}} \\
\cmidrule(lr){1-5}
\cmidrule(lr){6-10}
\cmidrule(lr){11-15}
\multicolumn{3}{c}{\textbf{Transition}}
& \textbf{Count} & \textbf{\%} &
\multicolumn{3}{c}{\textbf{Transition}}
& \textbf{Count} & \textbf{\%} &
\multicolumn{3}{c}{\textbf{Transition}}
& \textbf{Count} & \textbf{\%} \\
\midrule

fear & $\rightarrow$ & fear & 720 & 4.83 &
self-doubt & $\rightarrow$ & self-doubt & 214 & 5.11 &
self-doubt & $\rightarrow$ & self-doubt & 1,757 & 3.48 \\

self-doubt & $\rightarrow$ & self-doubt & 691 & 4.64 &
fear & $\rightarrow$ & fear & 155 & 3.70 &
fear & $\rightarrow$ & fear & 1,537 & 3.04 \\

hopeless & $\rightarrow$ & hopeless & 541 & 3.63 &
anxiety & $\rightarrow$ & anxiety & 124 & 2.96 &
anxiety & $\rightarrow$ & anxiety & 1,135 & 2.25 \\

anxiety & $\rightarrow$ & anxiety & 510 & 3.42 &
fear & $\rightarrow$ & self-doubt & 95 & 2.27 &
hopeful & $\rightarrow$ & hopeful & 1,117 & 2.21 \\

overwhelm & $\rightarrow$ & overwhelm & 456 & 3.06 &
self-doubt & $\rightarrow$ & fear & 91 & 2.17 &
loneliness & $\rightarrow$ & loneliness & 896 & 1.77 \\

\bottomrule
\end{tabular}%
}

\vspace{6pt}

% Additional corpus: PsyChain
{\color{revisionblue}
\begin{tabular}{@{}l c l r r@{}}
\toprule
\multicolumn{5}{c}{
\textbf{PsyChain (Synthetic)}
} \\
\cmidrule(lr){1-5}
\multicolumn{3}{c}{\textbf{Transition}}
& \textbf{Count} & \textbf{\%} \\
\midrule

self-doubt & $\rightarrow$ & self-doubt & 1,108 & 6.06 \\
fear & $\rightarrow$ & fear & 1,035 & 5.66 \\
self-aware & $\rightarrow$ & self-aware & 604 & 3.30 \\
self-doubt & $\rightarrow$ & fear & 585 & 3.20 \\
fear & $\rightarrow$ & self-doubt & 553 & 3.02 \\

\bottomrule
\end{tabular}
}
\end{table}

\subsection{Recovery Timing}
\begin{figure}[h]
    \centering
    \includegraphics[width=0.62\textwidth]{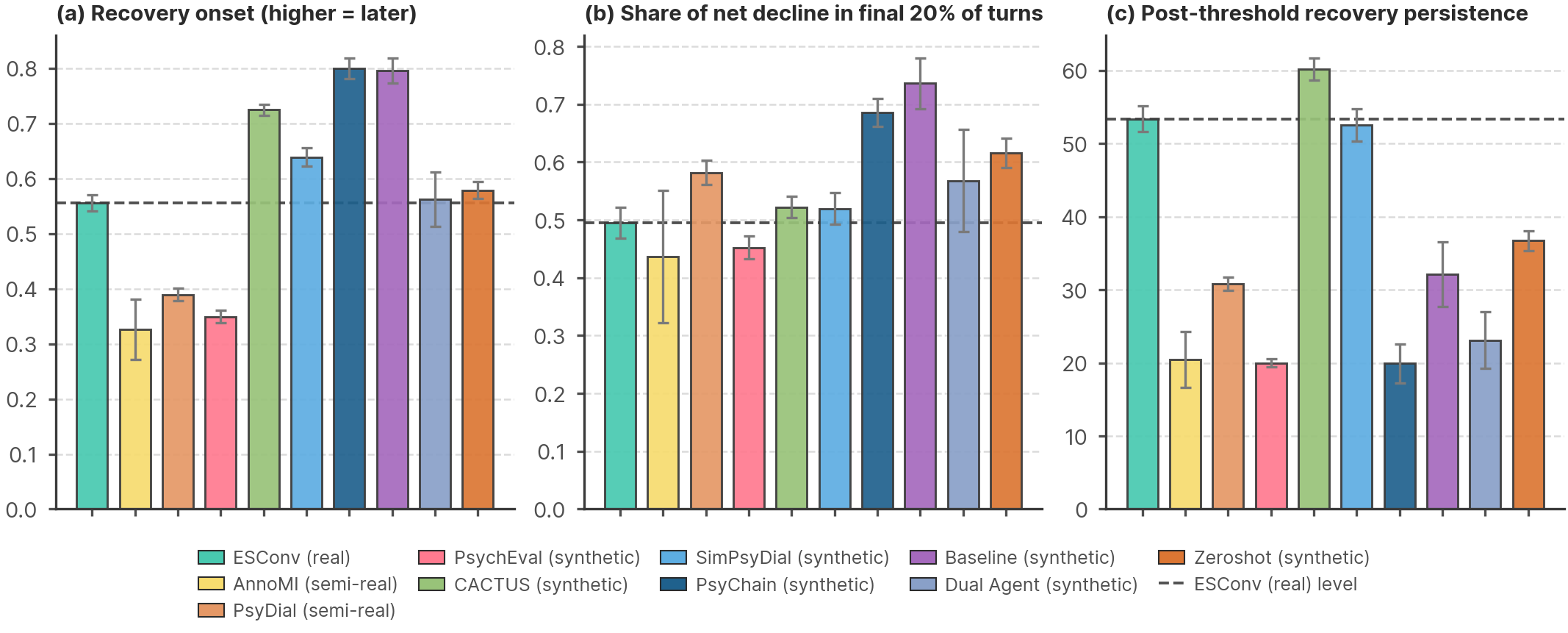}
    \captionsetup{font=footnotesize,skip=3pt}
    \caption{
    Recovery timing and consolidation. \textcolor{revisionblue}{PsyChain included.}
    \textbf{(a)} Onset: normalized position of the first turn with
    distress $\leq 0.33$ (higher is later).
    \textbf{(b)} Last-mile share: fraction of net decline in the final
    20\% of turns.
    \textbf{(c)} Persistence: fraction of post-onset turns that remain
    recovered. Dashed lines mark ESConv.
    }
    \label{fig:scripted_recovery}
\end{figure}

\subsection{Counselor Strategy}
\begin{figure}[H]
    \centering
    \includegraphics[width=\textwidth]{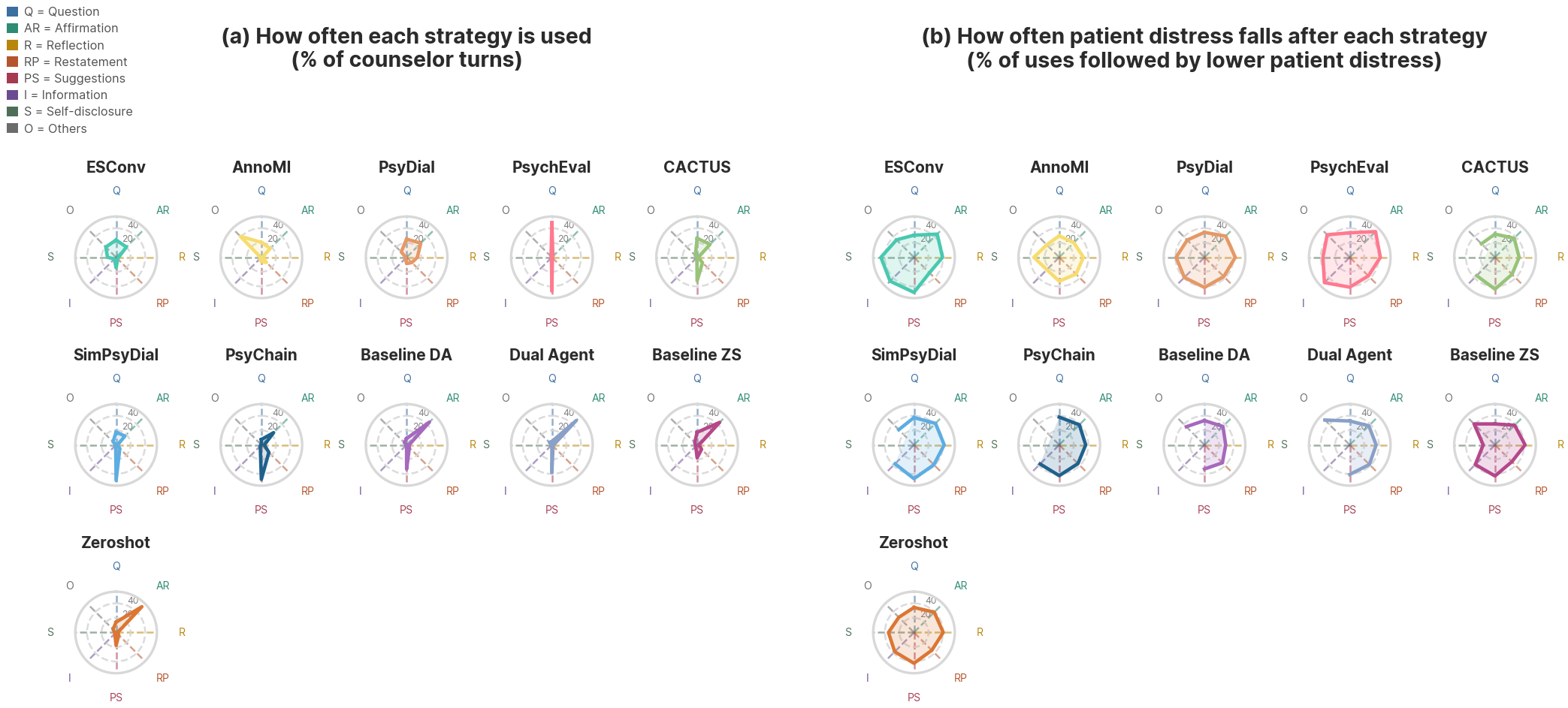}
    \caption{
    Counselor strategy use across datasets. \textcolor{revisionblue}{PsyChain included.}
    \textbf{(a)} Percentage of counselor turns assigned to each strategy.
    \textbf{(b)} Percentage of each strategy's uses followed by lower
    patient distress. All radar plots share the same axes; abbreviations
    are defined in the legend. %\msnote{This might be more interpretable as just a table of percents and would save space}
    }
    \label{fig:strategy_pairs}
\end{figure}

\subsection{Archetype Distributions}

\begin{table}[H]
    \centering
    \captionsetup{font=footnotesize,skip=2pt}

    \caption{
    Distress-trajectory archetype distribution by dataset
    (\% of conversations).
    }
    \label{tab:trajectory_patterns}

    \resizebox{0.78\linewidth}{!}{%
    \begin{tabular}{lccccc}
    \toprule[1.2pt]
    \textbf{Dataset}
    & \shortstack{\textbf{Early}\\\textbf{resolution}}
    & \shortstack{\textbf{Late}\\\textbf{recovery}}
    & \shortstack{\textbf{Persistent}\\\textbf{moderate}\\\textbf{distress}}
    & \shortstack{\textbf{Steady}\\\textbf{de-escalation}\\\textbf{(high distress)}}
    & \shortstack{\textbf{Unresolved}\\\textbf{high}\\\textbf{distress}} \\
    \midrule

    \rowcolor{gray!15}
    \multicolumn{6}{l}{\textbf{\textit{Real}}} \\
    ESConv      & 30  & 27   & 16  & 15 & 12 \\

    \midrule
    \rowcolor{gray!15}
    \multicolumn{6}{l}{\textbf{\textit{Semi-real}}} \\
    AnnoMI      & 5   & 5    & 64  & 7  & 18 \\
    PsyDial     & 15  & 24   & 21  & 27 & 13 \\

    \midrule
    \rowcolor{gray!15}
    \multicolumn{6}{l}{\textbf{\textit{Synthetic}}} \\

    \textcolor{revisionblue}{PsyChain}
    & \textcolor{revisionblue}{2}
    & \textcolor{revisionblue}{17}
    & \textcolor{revisionblue}{1}
    & \textcolor{revisionblue}{57}
    & \textcolor{revisionblue}{22} \\

    PsychEval   & 8   & 10   & 56  & 14 & 12 \\
    CACTUS      & 25  & 34   & 2   & 27 & 12 \\
    SimPsyDial  & 24  & 29   & 14  & 26 & 7  \\
    Baseline DA & 1.2 & 1.5  & 0.0 & 4  & 93 \\
    Dual Agent  & 3   & 5    & 2   & 11 & 78 \\
    Baseline ZS & 0.3 & 15   & 0.0 & 39 & 45 \\
    Zero-shot   & 13  & 18   & 4   & 45 & 20 \\

    \bottomrule[1.2pt]
    \end{tabular}%
    }
\end{table}

\section{Prompts-tuning Experiments}
\label{app:tuned_prompts}

\subsection{Recovery-Dynamics Prompt}
\label{app:prompt_recovery}
This intervention modifies the zero-shot generation prompt to reduce
stage-driven recovery and encourage gradual, sustained improvement that follows
from the interaction rather than from conversation position.

\begin{promptbox}{Recovery-Dynamics Tuning}
\textbf{Prompt addition:}

RECOVERY DYNAMICS:

The patient should not recover simply because the conversation is progressing
or approaching its end. Emotional change should follow from what has actually
happened in the interaction.

EARLY AND MIDDLE CONVERSATION:
During the earlier part of the conversation, helpful counselor responses may
make the patient more willing to elaborate, more reflective, or slightly less
overwhelmed, but these changes should usually be small. Do not treat a few
supportive responses as sufficient for substantial emotional recovery.

Allow the patient's core distress to remain present while the problem is still
being explored. Meaningful recovery should generally require sustained
understanding, several relevant counselor responses, or the development of a
concrete new perspective or coping direction.

RECOVERY:
When improvement begins, let it accumulate across several turns rather than
appearing as a sudden transition. Small improvements may occur earlier, but
substantial relief should emerge only after enough of the underlying concern
has been explored.

Once meaningful relief is reached, preserve some of that progress unless the
conversation introduces a genuine reason for renewed distress. Avoid arbitrary
relapse for the sake of emotional variety.

ENDING:
Do not force the patient to become positive, hopeful, or fully resolved by the
end. A realistic conversation may end with partial relief, greater clarity,
or willingness to cope while substantial distress remains.

The final emotional state should reflect how much progress was actually made
during the interaction. Do not add a final positive turn solely to provide
closure.

Overall, prefer partial and gradual improvement over either:
\begin{enumerate}[nosep,leftmargin=*]
\item remaining emotionally unchanged for most of the conversation, or
\item abruptly becoming positive near the end.
\end{enumerate}
\end{promptbox}

\subsection{Archetype-diversity prompt}
\label{app:prompt_diversity}
This intervention augments the original zero-shot prompt with one or two
sampled trajectory conditions per conversation. The base generation prompt,
including its message-numbered schedule, is otherwise unchanged.

\begin{promptbox}{Archetype-Diversity Tuning}
\textbf{Prompt addition:}

TRAJECTORY CONDITIONS:

The patient's emotional course should follow from the interaction and the
following starting conditions:

\{\{sampled\_conditions\}\}

These conditions describe the difficulty and constraints of the situation,
not a required trajectory or ending. Do not force the conversation into a
predefined recovery pattern.
\end{promptbox}

For each conversation, \texttt{\{\{sampled\_conditions\}\}} contains one or two
conditions sampled from different categories, with one of two possible values
selected for each category:

\begin{itemize}[nosep,leftmargin=*]
    \item \textbf{Problem resolvability}
    \begin{itemize}[nosep,leftmargin=*]
        \item The core problem can be meaningfully addressed within this one
        conversation.
        \item The core problem cannot be resolved within one conversation.
    \end{itemize}

    \item \textbf{Coping resources}
    \begin{itemize}[nosep,leftmargin=*]
        \item The patient has existing coping resources they can draw on.
        \item The patient has few existing coping resources.
    \end{itemize}

    \item \textbf{Receptiveness}
    \begin{itemize}[nosep,leftmargin=*]
        \item The patient is initially receptive to support.
        \item The patient is initially guarded.
    \end{itemize}

    \item \textbf{External stressor}
    \begin{itemize}[nosep,leftmargin=*]
        \item An external stressor persists and keeps affecting the patient
        during the conversation.
        \item No external stressor is actively intruding during the conversation.
    \end{itemize}

    \item \textbf{Improvement extent}
    \begin{itemize}[nosep,leftmargin=*]
        \item If improvement happens, it is likely to be partial.
        \item If improvement happens, it can be substantial.
    \end{itemize}

    \item \textbf{Setback likelihood}
    \begin{itemize}[nosep,leftmargin=*]
        \item Setbacks are likely after progress.
        \item Progress, once made, tends to hold.
    \end{itemize}
\end{itemize}

\subsection{Strategy-diversity prompt}
\label{app:prompt_strategy}
This intervention augments the original zero-shot prompt with an explicit
strategy-balance instruction for the counselor. All other generation
instructions, including the message-numbered schedule, remain unchanged.

\begin{promptbox}{Strategy-Diversity Tuning}
\textbf{Prompt addition:}

COUNSELOR STRATEGY BALANCE:

Do not default to validation or reassurance on every turn. Choose the response
type that best fits the immediate conversational need. Across the conversation,
use a varied mixture of:
\begin{itemize}[nosep,leftmargin=*]
    \item questions,
    \item reflection of feelings,
    \item restatement or paraphrasing,
    \item information,
    \item suggestions,
    \item affirmation/reassurance.
\end{itemize}

Use affirmation when it is specifically warranted rather than as a routine
opening to each response.
\end{promptbox}
The block is inserted in the middle-section counselor instructions immediately
after the instruction to maintain a steady, calm presence and validate the
patient's feelings naturally.

\subsection{Strategy--emotion coupling prompt}
\label{app:prompt_coupling}
This intervention augments the original zero-shot prompt with an instruction
that conditions the patient's emotional response on the affective quality of
the counselor's immediately preceding turn. All other generation instructions,
including the message-numbered schedule, remain unchanged. The intervention
targets counselor--patient coupling through the emotion channel and does not
explicitly refer to counselor strategy labels.

\begin{promptbox}{Strategy--Emotion Coupling Tuning}
\textbf{Prompt addition:}

AFFECTIVE CONTINGENCY:

The patient's next emotional state should respond to the emotional quality of
the counselor's immediately preceding message. Warm, well-attuned responses
may slightly reduce distress or increase openness; emotionally mismatched,
detached, or overly formulaic responses may leave distress unchanged or
increase it.

Do not follow a predetermined emotional arc independent of the counselor's
turn.
\end{promptbox}

\subsection{Turn-level relief prompt}
\label{app:prompt_turn_relief}
This intervention augments the original zero-shot prompt with an instruction
that makes the patient's immediate emotional response depend on how well the
counselor's preceding message addresses the patient's current concern. All other
generation instructions, including the message-numbered schedule, remain
unchanged.

\begin{promptbox}{Turn-Level Relief Tuning}
\textbf{Prompt addition:}

TURN-LEVEL CONTINGENCY:

For every patient response, first consider the counselor's immediately preceding
message. The patient's emotional reaction should reflect how well that response
fits their current concern.

A response that accurately addresses the patient's specific feeling or problem
may produce a small reduction in distress. A generic, repetitive, mistimed, or
unhelpful response should produce little improvement and may increase
frustration.

Do not improve simply because the conversation is approaching its end.
\end{promptbox}

\subsection{Prompt-tuning results (GPT 4o)}

\begin{figure}[H]
    \centering
    \includegraphics[width=\linewidth]{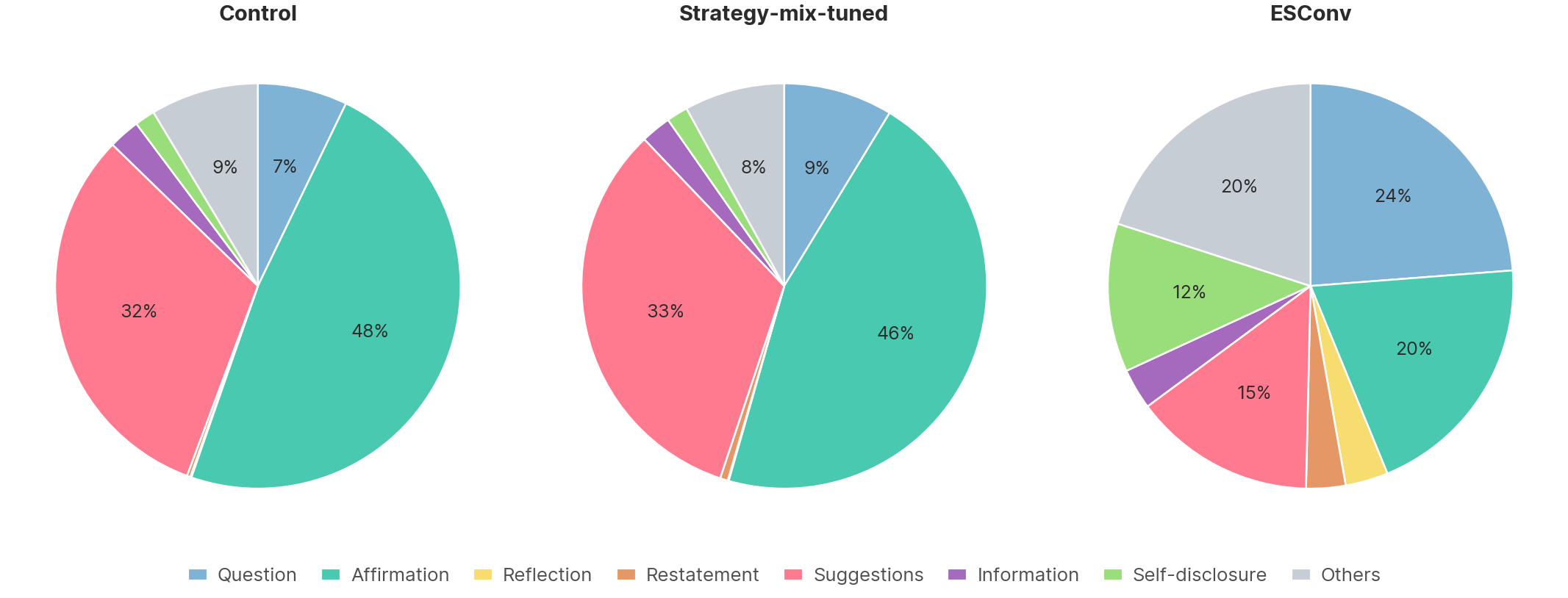}
    \caption{
    Strategy distributions for the control, strategy-mix-tuned, and ESConv conversations.
    }
    \label{fig:p5_strategy_pies}
\end{figure}

\begingroup
\color{revisionblue}

% ============================================================
% Evaluation criteria
% ============================================================

% ============================================================
% GPT-6.1 Sol
% ============================================================

\subsection{Additional Prompt Tuning Experiment:
\textbf{GPT-6.1 Sol}}

\begin{table}[H]
\color{revisionblue}
\captionsetup{
    labelfont={color=revisionblue},
    textfont={color=revisionblue}
}
\centering
\caption{Prompt-tuning results for GPT-6.1 Sol.}
\label{tab:sol_prompt_tuning}
\small
\setlength{\tabcolsep}{4pt}

\begin{tabular}{@{}llcccc@{}}
\toprule
\textbf{Measure} & \textbf{Metric}
& \textbf{Control}
& \textbf{Tuned}
& \textbf{ESConv}
& \textbf{Result} \\
\midrule

\rowcolor{gray!15}
\multicolumn{6}{@{}l}{
\textbf{\textit{Tier 1: varies with protocol}}} \\

\multirow{4}{*}{\blackcirc{1} Recovery}
 & End positive (\%)
 & 70.0 & 71.0 & 67.3 & \xmark \\
 & Recovery onset
 & 0.59 & 0.43 & 0.56 & \xmark \\
 & Last-mile share
 & 0.81 & 0.63 & 0.50 & \cmark \\
 & Post-onset persistence (\%)
 & 16.8 & 19.7 & 53.4 & \xmark \\
\addlinespace[2pt]

\multirow{2}{*}{\blackcirc{2}\ Archetype coverage}
 & JSD to ESConv
 & 0.534 & 0.497 & 0.00 & \xmark \\
 & Dominant archetype (\%)
 & 72.0 & 60.0 & 30.0 & \cmark$^{\dagger}$ \\
\midrule

\rowcolor{gray!15}
\multicolumn{6}{@{}l}{
\textbf{\textit{Tier 2: persists across prompt interventions}}} \\

\multirow{2}{*}{\blackcirc{3}\ Strategy mix}
 & Strategy evenness ($J$)
 & 0.73 & 0.72 & 0.89 & \xmark \\
 & Top-2 strategy share (\%)
 & 66.6 & 65.6 & 34.6 & \xmark \\

\blackcirc{4}\ Strategy-channel coupling
 & Stage-adjusted CMI (bits)
 & 0.010 & 0.009 & 0.039 & \xmark \\

\blackcirc{5}\ Emotion-channel coupling
 & Stage-adjusted CMI (bits)
 & 0.001 & 0.002 & 0.019 & \xmark \\

\multirow{2}{*}{\raisebox{0.8ex}{
\blackcirc{6} Turn-level relief}}
 & Strategy-explained $\Delta d$ (\%)
 & 1.7 & 1.1 & 1.9 & \xmark \\
 & Mean $\Delta d$ per counselor turn
 & $-0.009$ & $-0.009$ & $-0.041$ & \xmark \\

\bottomrule
\end{tabular}
\end{table}

% ============================================================
% Claude Sonnet 5.5
% ============================================================

\subsection{Additional Prompt Tuning Experiment:
\textbf{Claude Sonnet 5.5}}

\begin{table}[H]
\color{revisionblue}
\captionsetup{
    labelfont={color=revisionblue},
    textfont={color=revisionblue}
}
\centering
\caption{Prompt-tuning results for Claude Sonnet 5.5.}
\label{tab:claude_prompt_tuning}
\small
\setlength{\tabcolsep}{4pt}

\begin{tabular}{@{}llcccc@{}}
\toprule
\textbf{Measure} & \textbf{Metric}
& \textbf{Control}
& \textbf{Tuned}
& \textbf{ESConv}
& \textbf{Result} \\
\midrule

\rowcolor{gray!15}
\multicolumn{6}{@{}l}{
\textbf{\textit{Tier 1: varies with protocol}}} \\

\multirow{4}{*}{\blackcirc{1} Recovery}
 & End positive (\%)
 & 64.0 & 71.0 & 67.3 & \xmark \\
 & Recovery onset
 & 0.42 & 0.44 & 0.56 & \xmark \\
 & Last-mile share
 & 0.64 & 0.51 & 0.50 & \cmark \\
 & Post-onset persistence (\%)
 & 19.6 & 27.7 & 53.4 & \cmark$^{\dagger}$ \\
\addlinespace[2pt]

\multirow{2}{*}{\blackcirc{2}\ Archetype coverage}
 & JSD to ESConv
 & 0.304 & 0.197 & 0.00 & \cmark \\
 & Dominant archetype (\%)
 & 56.0 & 51.0 & 30.0 & \cmark$^{\dagger}$ \\
\midrule

\rowcolor{gray!15}
\multicolumn{6}{@{}l}{
\textbf{\textit{Tier 2: persists across prompt interventions}}} \\

\multirow{2}{*}{\blackcirc{3}\ Strategy mix}
 & Strategy evenness ($J$)
 & 0.70 & 0.73 & 0.89 & \cmark$^{\dagger}$ \\
 & Top-2 strategy share (\%)
 & 75.3 & 70.0 & 34.6 & \xmark \\

\blackcirc{4}\ Strategy-channel coupling
 & Stage-adjusted CMI (bits)
 & 0.009 & 0.012 & 0.039 & \xmark \\

\blackcirc{5}\ Emotion-channel coupling
 & Stage-adjusted CMI (bits)
 & 0.005 & 0.002 & 0.019 & \xmark \\

\multirow{2}{*}{\raisebox{0.8ex}{
\blackcirc{6} Turn-level relief}}
 & Strategy-explained $\Delta d$ (\%)
 & 0.8 & 0.8 & 1.9 & \xmark \\
 & Mean $\Delta d$ per counselor turn
 & $-0.012$ & $-0.012$ & $-0.041$ & \xmark \\

\bottomrule
\end{tabular}
\end{table}

% ============================================================
% Gemini 3.1 Pro
% ============================================================

\subsection{Additional Prompt Tuning Experiment:
\textbf{Gemini 3.1 Pro}}

\begin{table}[H]
\color{revisionblue}
\captionsetup{
    labelfont={color=revisionblue},
    textfont={color=revisionblue}
}
\centering
\caption{Prompt-tuning results for Gemini 3.1 Pro.}
\label{tab:gemini_prompt_tuning}
\small
\setlength{\tabcolsep}{4pt}

\begin{tabular}{@{}llcccc@{}}
\toprule
\textbf{Measure} & \textbf{Metric}
& \textbf{Control}
& \textbf{Tuned}
& \textbf{ESConv}
& \textbf{Result} \\
\midrule

\rowcolor{gray!15}
\multicolumn{6}{@{}l}{
\textbf{\textit{Tier 1: varies with protocol}}} \\

\multirow{4}{*}{\blackcirc{1} Recovery}
 & End positive (\%)
 & 44.0 & 51.0 & 67.3 & \cmark \\
 & Recovery onset
 & 0.52 & 0.49 & 0.56 & \xmark \\
 & Last-mile share
 & 0.62 & 0.52 & 0.50 & \cmark \\
 & Post-onset persistence (\%)
 & 24.0 & 27.1 & 53.4 & \xmark \\
\addlinespace[2pt]

\multirow{2}{*}{\blackcirc{2}\ Archetype coverage}
 & JSD to ESConv
 & 0.348 & 0.265 & 0.00 & \cmark$^{\dagger}$ \\
 & Dominant archetype (\%)
 & 51.0 & 50.0 & 30.0 & \xmark \\
\midrule

\rowcolor{gray!15}
\multicolumn{6}{@{}l}{
\textbf{\textit{Tier 2: persists across prompt interventions}}} \\

\multirow{2}{*}{\blackcirc{3}\ Strategy mix}
 & Strategy evenness ($J$)
 & 0.65 & 0.71 & 0.89 & \cmark$^{\dagger}$ \\
 & Top-2 strategy share (\%)
 & 71.9 & 68.5 & 34.6 & \xmark \\

\blackcirc{4}\ Strategy-channel coupling
 & Stage-adjusted CMI (bits)
 & 0.011 & 0.008 & 0.039 & \xmark \\

\blackcirc{5}\ Emotion-channel coupling
 & Stage-adjusted CMI (bits)
 & 0.009 & 0.008 & 0.019 & \xmark \\

\multirow{2}{*}{\raisebox{0.8ex}{
\blackcirc{6} Turn-level relief}}
 & Strategy-explained $\Delta d$ (\%)
 & 1.0 & 0.4 & 1.9 & \xmark \\
 & Mean $\Delta d$ per counselor turn
 & $-0.009$ & $-0.009$ & $-0.041$ & \xmark \\

\bottomrule
\end{tabular}
\end{table}

\section{Statistical Significance of Cross-Corpus Differences}
\label{app:cross_corpus_significance}
We compare each of the ten corpora with ESConv using conversation-level permutation tests, preserving within-conversation dependencies. We report Holm-adjusted $p$-values using thresholds
$p_{\mathrm{Holm}}<0.001$, $<0.01$, and $<0.05$;
\emph{n.s.} denotes $p_{\mathrm{Holm}}\geq 0.05$.
Thresholds are applied to adjusted $p$-values. Holm correction is applied separately to each prespecified family. Trajectory and diversity analyses use 20,000 permutations and downstream-effect analyses use 100,000, so that the smallest attainable adjusted $p$-value lies below 0.001. Coupling uses 1,000 outer permutations (with 100 inner permutations) because each permutation recomputes the bias correction; its smallest attainable adjusted $p$-value is therefore about 0.01.

\subsection{Distress Trajectories (Figure~3a)}
\label{app:significance_trajectory}
We test the root-mean-square (RMS) distance between corpus-level mean distress trajectories on a 20-point grid. We also test shape-only RMS distance after centering each mean trajectory and the signed difference in overall mean distress. All ten corpora differ from ESConv in both RMS distance and shape-only RMS distance (Holm-adjusted $p<0.001$). The overall mean distress level for AnnoMI is not significantly different ($p_{\mathrm{Holm}}\geq 0.05$), demonstrating that trajectory shape differences need not be explained by average distress levels alone.
\begin{table}[H]
\color{revisionblue}
\captionsetup{
    labelfont={color=revisionblue},
    textfont={color=revisionblue}
}

\caption{
Distress-trajectory comparisons against ESConv.
Mean-level $\Delta$ is corpus minus ESConv;
bold entries indicate $p_{\mathrm{Holm}}<0.05$;
n.s.\ indicates nonsignificance.
Each statistic is corrected across ten corpus comparisons.
}

\label{tab:significance_trajectory}
\centering
\scriptsize
\setlength{\tabcolsep}{3pt}
\resizebox{\linewidth}{!}{%
\begin{tabular}{@{}lcccccc@{}}
\toprule
Corpus & RMS distance & $p_{\mathrm{Holm}}$ & Shape-only RMS & $p_{\mathrm{Holm}}$ & Mean-level $\Delta$ & $p_{\mathrm{Holm}}$ \\
\midrule
AnnoMI & 0.1189 & \textbf{$<0.001$} & 0.1175 & \textbf{$<0.001$} & -0.0183 & $\mathrm{n.s.}$ \\
PsyDial & 0.0575 & \textbf{$<0.001$} & 0.0521 & \textbf{$<0.001$} & 0.0244 & \textbf{$<0.001$} \\
PsychEval & 0.0960 & \textbf{$<0.001$} & 0.0951 & \textbf{$<0.001$} & -0.0124 & \textbf{$<0.001$} \\
CACTUS & 0.0592 & \textbf{$<0.001$} & 0.0382 & \textbf{$<0.001$} & 0.0453 & \textbf{$<0.001$} \\
SimPsyDial & 0.0355 & \textbf{$<0.001$} & 0.0221 & \textbf{$<0.001$} & 0.0278 & \textbf{$<0.001$} \\
\rowcolor{revisionblue!10}
PsyChain & 0.1542 & $<0.001$ & 0.0661 & $<0.001$ & 0.1393 & $<0.001$ \\
Baseline DA & 0.2648 & \textbf{$<0.001$} & 0.1401 & \textbf{$<0.001$} & 0.2246 & \textbf{$<0.001$} \\
Dual Agent & 0.2062 & \textbf{$<0.001$} & 0.1073 & \textbf{$<0.001$} & 0.1761 & \textbf{$<0.001$} \\
Baseline ZS & 0.1906 & \textbf{$<0.001$} & 0.0857 & \textbf{$<0.001$} & 0.1702 & \textbf{$<0.001$} \\
Zeroshot & 0.1021 & \textbf{$<0.001$} & 0.0423 & \textbf{$<0.001$} & 0.0930 & \textbf{$<0.001$} \\
\bottomrule
\end{tabular}%
}

\end{table}

\subsection{Strategy-Specific Downstream Distress (Figure~3b)}
\label{app:significance_strategy_effects}

For each eligible corpus--strategy pair, we test the
difference in mean subsequent patient distress change,
$\Delta = \overline{\Delta d}_{\mathrm{corpus}}
-\overline{\Delta d}_{\mathrm{ESConv}}$.
Affirmation and Reassurance differs significantly in
nine of ten corpora after Holm correction.
PsyChain exhibits significant differences in Question,
Affirmation and Reassurance, Providing Suggestions,
and Others.
The matrix reports adjusted $p$-value thresholds for
all eligible tests; dashes indicate fewer than
20 observations.

\begin{table}[H]
\color{revisionblue}
\captionsetup{
    labelfont={color=revisionblue},
    textfont={color=revisionblue}
}
\caption{
Holm-adjusted $p$-values for corpus--ESConv
differences in downstream distress change by strategy
(73-test family).
Bold entries indicate $p_{\mathrm{Holm}}<0.05$;
-- indicates insufficient observations ($n<20$).
Abbreviations follow the strategy names in the main text.
}
\label{tab:significance_strategy_effects}

\centering
\scriptsize
\setlength{\tabcolsep}{3pt}
\resizebox{\linewidth}{!}{%
\begin{tabular}{@{}lcccccccc@{}}
\toprule
Corpus
& Ques.
& Affirm.
& Reflect.
& Restate
& Suggest.
& Info.
& Self-disc.
& Other \\
\midrule

AnnoMI
& $\mathrm{n.s.}$
& \textbf{$<0.05$}
& $\mathrm{n.s.}$
& $\mathrm{n.s.}$
& $\mathrm{n.s.}$
& $\mathrm{n.s.}$
& $\mathrm{n.s.}$
& $\mathrm{n.s.}$ \\

PsyDial
& $\mathrm{n.s.}$
& \textbf{$<0.01$}
& $\mathrm{n.s.}$
& $\mathrm{n.s.}$
& \textbf{$<0.01$}
& \textbf{$<0.05$}
& \textbf{$<0.01$}
& $\mathrm{n.s.}$ \\

PsychEval
& \textbf{$<0.01$}
& $\mathrm{n.s.}$
& $\mathrm{n.s.}$
& $\mathrm{n.s.}$
& \textbf{$<0.01$}
& $\mathrm{n.s.}$
& \textbf{$<0.01$}
& \textbf{$<0.01$} \\

CACTUS
& $\mathrm{n.s.}$
& \textbf{$<0.01$}
& $\mathrm{n.s.}$
& $\mathrm{n.s.}$
& $\mathrm{n.s.}$
& $\mathrm{n.s.}$
& --
& $\mathrm{n.s.}$ \\

SimPsyDial
& \textbf{$<0.05$}
& \textbf{$<0.01$}
& $\mathrm{n.s.}$
& $\mathrm{n.s.}$
& $\mathrm{n.s.}$
& $\mathrm{n.s.}$
& --
& \textbf{$<0.01$} \\
\rowcolor{revisionblue!10}
\textcolor{revisionblue}{PsyChain}
& \textbf{$<0.01$}
& \textbf{$<0.01$}
& $\mathrm{n.s.}$
& $\mathrm{n.s.}$
& \textbf{$<0.01$}
& $\mathrm{n.s.}$
& --
& \textbf{$<0.01$} \\

Baseline DA
& $\mathrm{n.s.}$
& \textbf{$<0.01$}
& $\mathrm{n.s.}$
& $\mathrm{n.s.}$
& \textbf{$<0.01$}
& --
& --
& $\mathrm{n.s.}$ \\

Dual Agent
& $\mathrm{n.s.}$
& \textbf{$<0.01$}
& $\mathrm{n.s.}$
& $\mathrm{n.s.}$
& \textbf{$<0.01$}
& --
& --
& $\mathrm{n.s.}$ \\

Baseline ZS
& $\mathrm{n.s.}$
& \textbf{$<0.01$}
& $\mathrm{n.s.}$
& $\mathrm{n.s.}$
& $\mathrm{n.s.}$
& $\mathrm{n.s.}$
& \textbf{$<0.01$}
& $\mathrm{n.s.}$ \\

Zeroshot
& $\mathrm{n.s.}$
& \textbf{$<0.01$}
& $\mathrm{n.s.}$
& $\mathrm{n.s.}$
& \textbf{$<0.01$}
& $\mathrm{n.s.}$
& \textbf{$<0.01$}
& \textbf{$<0.01$} \\

\bottomrule
\end{tabular}%
}
\end{table}

\subsection{Stage-Adjusted Strategy and Emotion Coupling (Figure~3c)}
\label{app:significance_coupling}

We compare stage-adjusted, bias-corrected conditional mutual
information (CMI) against ESConv separately for the strategy
and emotion channels. Eight of ten corpora show significant
differences in each channel after Holm correction.
CACTUS and PsychEval do not significantly differ in the
strategy channel, while AnnoMI and PsyDial do not
significantly differ in the emotion channel.
PsyChain exhibits significantly weaker coupling in both
channels (strategy: 0.0040 vs.\ 0.0391 bits;
emotion: 0.0066 vs.\ 0.0192 bits).
Therefore, coupling discrepancies are widespread
but not universal.

\begin{table}[H]
\color{revisionblue}
\captionsetup{
    labelfont={color=revisionblue},
    textfont={color=revisionblue}
}
\caption{
Stage-adjusted, bias-corrected CMI (bits) and
Holm-adjusted $p$-value thresholds against ESConv.
Correction is performed separately across ten corpus
comparisons for each channel. Bold entries indicate
significance at $0.05$; n.s.\ indicates nonsignificance.
}
\label{tab:significance_coupling}

\centering
\small
\setlength{\tabcolsep}{4pt}
\begin{tabular}{@{}lcccc@{}}
\toprule
Corpus & Strategy CMI & $p_{\mathrm{Holm}}$
& Emotion CMI & $p_{\mathrm{Holm}}$ \\
\midrule

ESConv
& 0.0391 & --
& 0.0192 & -- \\

AnnoMI
& 0.0116 & \textbf{$<0.05$}
& 0.0107 & $\mathrm{n.s.}$ \\

PsyDial
& 0.0268 & \textbf{$<0.01$}
& 0.0195 & $\mathrm{n.s.}$ \\

PsychEval
& 0.0344 & $\mathrm{n.s.}$
& 0.0076 & \textbf{$<0.01$} \\

CACTUS
& 0.0399 & $\mathrm{n.s.}$
& 0.0094 & \textbf{$<0.01$} \\

SimPsyDial
& 0.0157 & \textbf{$<0.01$}
& 0.0040 & \textbf{$<0.01$} \\
\rowcolor{revisionblue!10}
\textcolor{revisionblue}{PsyChain}
& \textcolor{revisionblue}{0.0040}
& \textbf{\textcolor{revisionblue}{$<0.01$}}
& \textcolor{revisionblue}{0.0066}
& \textbf{\textcolor{revisionblue}{$<0.01$}} \\

Baseline DA
& 0.0034 & \textbf{$<0.01$}
& 0.0027 & \textbf{$<0.01$} \\

Dual Agent
& 0.0035 & \textbf{$<0.01$}
& 0.0016 & \textbf{$<0.01$} \\

Baseline ZS
& 0.0093 & \textbf{$<0.01$}
& 0.0006 & \textbf{$<0.01$} \\

Zeroshot
& 0.0143 & \textbf{$<0.01$}
& 0.0073 & \textbf{$<0.01$} \\

\bottomrule
\end{tabular}
\end{table}

\subsection{Inter-Conversation Trajectory Diversity (Figure~3d)}
\label{app:significance_diversity}

We compare the mean pairwise distance between smoothed
distress trajectories. All ten comparison corpora exhibit
significantly lower trajectory diversity than ESConv
(Holm-adjusted $p<0.001$). Although SimPsyDial is
numerically closest to ESConv, its difference remains
statistically significant. PsyChain also exhibits
substantially lower diversity than ESConv
(0.6591 vs.\ 1.0762; $p_{\mathrm{Holm}}<0.001$).

\begin{table}[H]
\color{revisionblue}
\captionsetup{
    labelfont={color=revisionblue},
    textfont={color=revisionblue}
}
\caption{
Inter-conversation trajectory diversity, measured by
mean pairwise distance between smoothed trajectories.
$\Delta$ is corpus minus ESConv.
Holm correction is applied across ten corpus comparisons.
}
\label{tab:significance_diversity}

\centering
\small
\setlength{\tabcolsep}{5pt}
\begin{tabular}{@{}lccc@{}}
\toprule
Corpus & Diversity & $\Delta$ vs. ESConv & $p_{\mathrm{Holm}}$ \\
\midrule
ESConv & 1.0762 & -- & -- \\
AnnoMI & 0.9336 & -0.1426 & \textbf{$<0.01$} \\
PsyDial & 0.9590 & -0.1172 & \textbf{$<0.01$} \\
PsychEval & 0.8438 & -0.2324 & \textbf{$<0.01$} \\
CACTUS & 0.9129 & -0.1634 & \textbf{$<0.01$} \\
SimPsyDial & 1.0099 & -0.0663 & \textbf{$<0.01$} \\
\rowcolor{revisionblue!10}
\textcolor{revisionblue}{PsyChain}
& \textcolor{revisionblue}{0.6591}
& \textcolor{revisionblue}{-0.4172}
& \textbf{\textcolor{revisionblue}{$<0.001$}} \\

Baseline DA & 0.5831 & -0.4931 & \textbf{$<0.01$} \\
Dual Agent & 0.7301 & -0.3461 & \textbf{$<0.01$} \\
Baseline ZS & 0.7149 & -0.3614 & \textbf{$<0.01$} \\
Zeroshot & 0.8113 & -0.2649 & \textbf{$<0.01$} \\
\bottomrule
\end{tabular}
\end{table}

\subsection{Significance tests for patient-level sentiment dynamics 
(Table~\ref{tab:micro_summary})}
\label{tab:micro_summary_significance}
We compare each corpus with ESConv for the seven metrics reported in Table~\ref{tab:micro_summary}. We retain the original descriptive values in Table~\ref{tab:micro_summary} and report the corresponding Holm-adjusted significance thresholds below. For conversation-level binary proportions (Start Neg\%, Stable\%, and End Pos\%), we use two-sided Fisher exact tests. For SCR, median first negative-to-positive transition turn, recovery-to-relapse ratio, and recovery-trigram percentage, we use conversation-level permutation tests (the ratio is tested on the log scale). Holm correction is applied separately across the nine corpus comparisons for each metric. The transition-turn analysis includes only conversations exhibiting a negative-to-positive flip. All reported p-values are adjusted; n.s. indicates $p_{\mathrm{Holm}}\geq0.05$.

\begin{table}[H]
\color{revisionblue}
\captionsetup{
    labelfont={color=revisionblue},
    textfont={color=revisionblue}
}
\caption{
Significance tests for Table~\ref{tab:micro_summary}
against ESConv. Entries are Holm-adjusted
$p$-value thresholds across 10 corpus comparisons
per metric; bold indicates $p_{\mathrm{Holm}}<0.05$,
and n.s.\ denotes nonsignificance.
}

\centering
\scriptsize
\setlength{\tabcolsep}{3pt}
\renewcommand{\arraystretch}{1.08}
\resizebox{\linewidth}{!}{%
\begin{tabular}{@{}lccccccc@{}}
\toprule
\textbf{Corpus}
& \textbf{Start Neg\%}
& \textbf{SCR}
& \textbf{Stable\%}
& \textbf{End Pos\%}
& \textbf{Turn}
& \textbf{Ratio}
& \textbf{Rec\%} \\
\midrule

\multicolumn{8}{@{}l}{\textit{Semi-real}} \\

AnnoMI
& \textbf{$<0.001$}
& \textit{n.s.}
& \textbf{$<0.01$}
& \textbf{$<0.001$}
& \textbf{$<0.01$}
& \textit{n.s.}
& \textbf{$<0.001$} \\

PsyDial
& \textbf{$<0.001$}
& \textbf{$<0.001$}
& \textit{n.s.}
& \textbf{$<0.001$}
& \textbf{$<0.001$}
& \textbf{$<0.001$}
& \textit{n.s.} \\

\midrule
\multicolumn{8}{@{}l}{\textit{Synthetic}} \\

Baseline DA
& \textbf{$<0.001$}
& \textbf{$<0.001$}
& \textbf{$<0.001$}
& \textbf{$<0.001$}
& \textbf{$<0.001$}
& \textbf{$<0.001$}
& \textbf{$<0.001$} \\

Baseline ZS
& \textbf{$<0.001$}
& \textbf{$<0.001$}
& \textbf{$<0.001$}
& \textbf{$<0.001$}
& \textbf{$<0.001$}
& \textit{n.s.}
& \textbf{$<0.001$} \\

CACTUS
& \textit{n.s.}
& \textbf{$<0.001$}
& \textbf{$<0.001$}
& \textbf{$<0.001$}
& \textbf{$<0.001$}
& \textit{n.s.}
& \textbf{$<0.001$} \\

Dual Agent
& \textit{n.s.}
& \textbf{$<0.001$}
& \textbf{$<0.001$}
& \textbf{$<0.001$}
& \textbf{$<0.001$}
& \textbf{$<0.001$}
& \textit{n.s.} \\

PsychEval
& \textbf{$<0.001$}
& \textbf{$<0.001$}
& \textbf{$<0.001$}
& \textbf{$<0.001$}
& \textit{n.s.}
& \textbf{$<0.001$}
& \textbf{$<0.001$} \\

SimPsyDial
& \textbf{$<0.05$}
& \textbf{$<0.001$}
& \textit{n.s.}
& \textbf{$<0.001$}
& \textit{n.s.}
& \textit{n.s.}
& \textbf{$<0.001$} \\

\rowcolor{revisionblue!10}
PsyChain
& \textit{n.s.}
& \textbf{$<0.001$}
& \textit{n.s.}
& \textbf{$<0.001$}
& \textbf{$<0.001$}
& \textbf{$<0.001$}
& \textbf{$<0.001$} \\

Zero-shot
& \textbf{$<0.001$}
& \textbf{$<0.001$}
& \textit{n.s.}
& \textbf{$<0.001$}
& \textbf{$<0.001$}
& \textbf{$<0.001$}
& \textit{n.s.} \\

\bottomrule
\end{tabular}%
}
\end{table}

\section{Emotion Annotation Reliability and Model Selection}
\label{app:emotion_annotation_reliability}

To validate the label quality and we independently annotated 50 ESConv conversations (705 client turns) using GPT-5.6 Luna (July 2026), Claude Opus 4.8 (May 2026), and Gemini 3.6 Flash (July 2026). We measured pairwise exact emotion-label agreement, Cohen's $\kappa$, emotion-polarity agreement and $\kappa$, Pearson correlation of derived distress scores, and agreement after treating synonymous emotion labels as equivalent. The results are presented in Table~\ref{tab:emotion_annotation_agreement}.

\begin{table}[H]
\color{revisionblue}
\captionsetup{
    labelfont={color=revisionblue},
    textfont={color=revisionblue}
}
\caption{Pairwise cross-model emotion-annotation agreement on 705 client turns from 50 ESConv conversations.}
\label{tab:emotion_annotation_agreement}
\centering
\scriptsize
\setlength{\tabcolsep}{3pt}
\resizebox{\linewidth}{!}{%
\begin{tabular}{@{}lrrrrrr@{}}
\toprule
Model pair & Exact (\%) & $\kappa$ & Polarity (\%) & Polarity $\kappa$ & Distress $r$ & Exact/syn. (\%) \\
\midrule
Luna -- Opus 4.8 & 69.9 & 0.68 & 87.5 & 0.79 & 0.85 & 80.9 \\
Luna -- Gemini 3.6 Flash & 67.2 & 0.64 & 81.8 & 0.72 & 0.84 & 75.9 \\
Opus 4.8 -- Gemini 3.6 Flash & 65.2 & 0.62 & 79.0 & 0.67 & 0.81 & 74.0 \\
\bottomrule
\end{tabular}%
}

\end{table}

We find that GPT-5.6 Luna shows substantial agreement with Opus 4.8 (69.9\% exact agreement; $\kappa=0.68$) and Gemini 3.6 Flash (67.2\%; $\kappa=0.64$), comparable to the agreement between Opus 4.8 and Gemini 3.6 Flash (65.2\%; $\kappa=0.62$). Agreement is higher for emotion polarity (87.5\% and 81.8\%, respectively), and the derived distress scores are strongly correlated ($r=0.85$ and $0.84$). When synonymous labels are counted as matches, agreement with Luna rises to 80.9\% and 75.9\%, respectively. 

However, Opus 4.8 is approximately 20 times more expensive than Luna, while Gemini 3.6 Flash is also 3 times more expensive. Given the large amount of utterances we label across all ten datasets, we selected GPT-5.6 Luna for the full annotation pipeline because it provides cross-model consistency at a practical inference cost and ensured label consistency across all datasets included.

\section{Interpolation Sensitivity}

\label{app:interpolation}

To test the robustness of our conclusions, we assess the sensitivity of our trajectory-based analyses
to interpolation resolution by resampling the original
turn-level distress sequences onto
$K\in\{20,30,40,50\}$ equally spaced progress points
using linear interpolation. The original experiments
use $K=20$. We examine three aspects of robustness:
\textbf{(i) mean distress trajectories and recovery trends,
(ii) inter-conversation trajectory diversity, and
(iii) trajectory archetype distributions.}
Metrics computed directly from the original turn
sequences, including recovery onset, last-mile share,
persistence, SCR, strategy trigrams, and
strategy--emotion coupling, do not depend on the
interpolation grid.

\paragraph{Mean distress trajectories.}
Figure~\ref{fig:grid_trajectory} compares mean distress
trajectories across interpolation resolutions.
The resulting trajectories remain highly consistent
with the original $K=20$ representation.
Across corpora, the minimum correlation with the
$K=20$ mean trajectory is $r=0.966$, $0.967$, and
$0.970$ for $K=30$, $40$, and $50$, respectively.
The corresponding maximum pointwise differences are
$0.025$, $0.023$, and $0.019$ on the normalized
$[0,1]$ distress scale.

We additionally calculate early-to-late distress
change as the difference between mean distress in
the second and first halves of each conversation.
Negative values indicate decreasing distress.
The corpus ranking under this metric remains
identical across all resolutions
(Kendall's $\tau=1.00$), indicating that the
relative recovery trends are insensitive to the
choice of interpolation grid.

\begin{figure}[H]
    \centering
    \captionsetup{
    labelfont={color=revisionblue},
    textfont={color=revisionblue}
}
    \includegraphics[
        width=0.75\linewidth,
        height=0.65\textheight,
        keepaspectratio
    ]{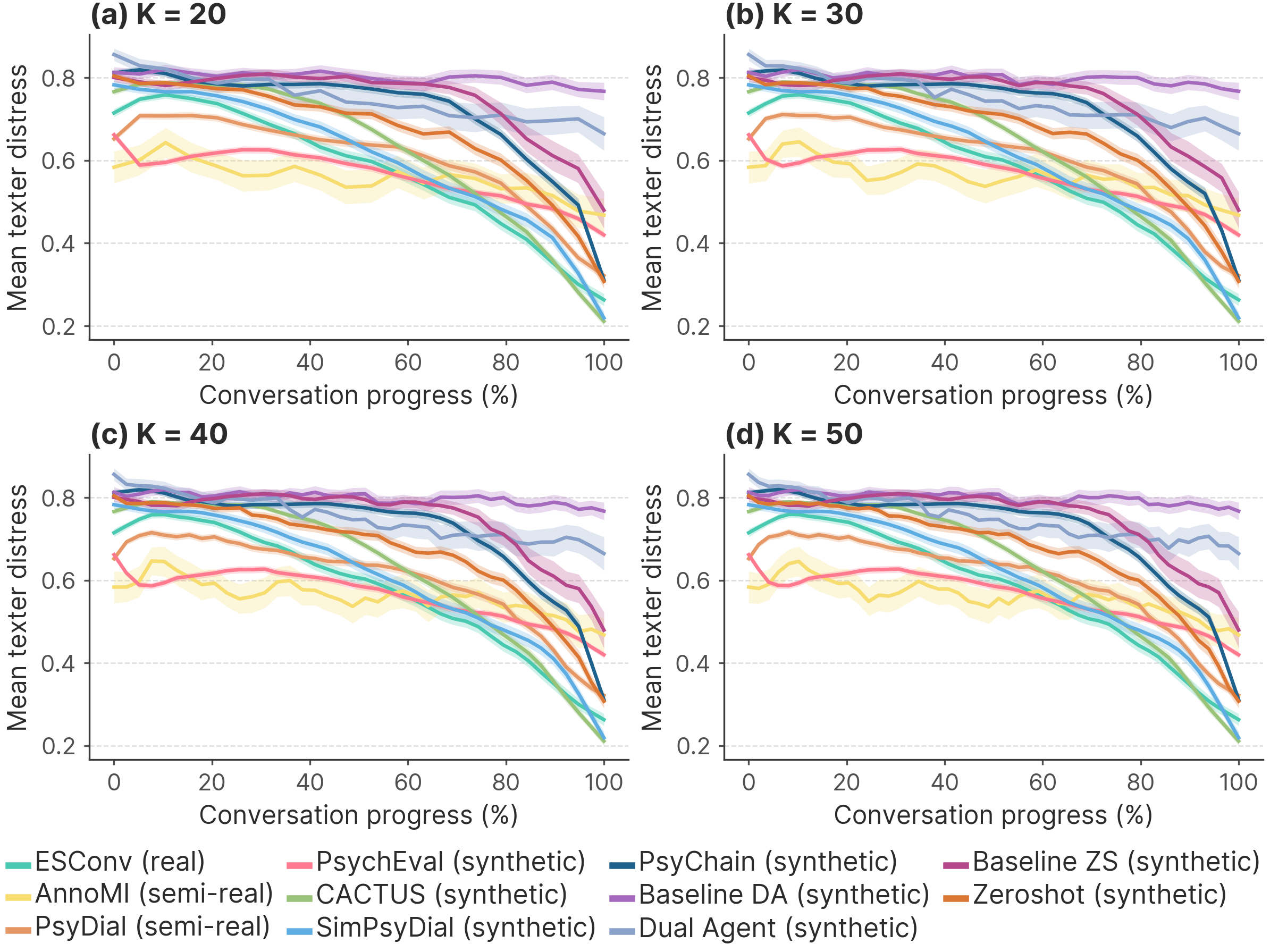}
    \caption{\textcolor{revisionblue}{
    Sensitivity of mean distress trajectories to
    interpolation resolution ($K=20,30,40,50$).
    The overall trajectory shapes and relative
    recovery patterns remain consistent across
    resolutions.}}
    \label{fig:grid_trajectory}
\end{figure}

\paragraph{Inter-conversation trajectory diversity.}
Figure~\ref{fig:grid_diversity} examines the
sensitivity of the diversity analysis in
Figure~3(d). Relative to $K=20$, Kendall's rank correlation
is $\tau=1.00$ at $K=30$ and $\tau=0.96$
at both $K=40$ and $K=50$.
ESConv remains the most diverse corpus across
all resolutions, followed by SimPsyDial,
while Baseline DA consistently exhibits the
lowest diversity. The only ranking change
at higher resolutions involves the
near-tied AnnoMI and CACTUS corpora,
whose normalized diversity scores differ
by less than $0.003$.
Thus, the principal inter-conversation diversity
findings remain robust under the original
distance definition and proportionally
scaled smoothing.

\begin{figure}[H]
    \centering
    \captionsetup{
    labelfont={color=revisionblue},
    textfont={color=revisionblue}
}
    \includegraphics[
        width=0.75\linewidth,
        height=0.65\textheight,
        keepaspectratio
    ]{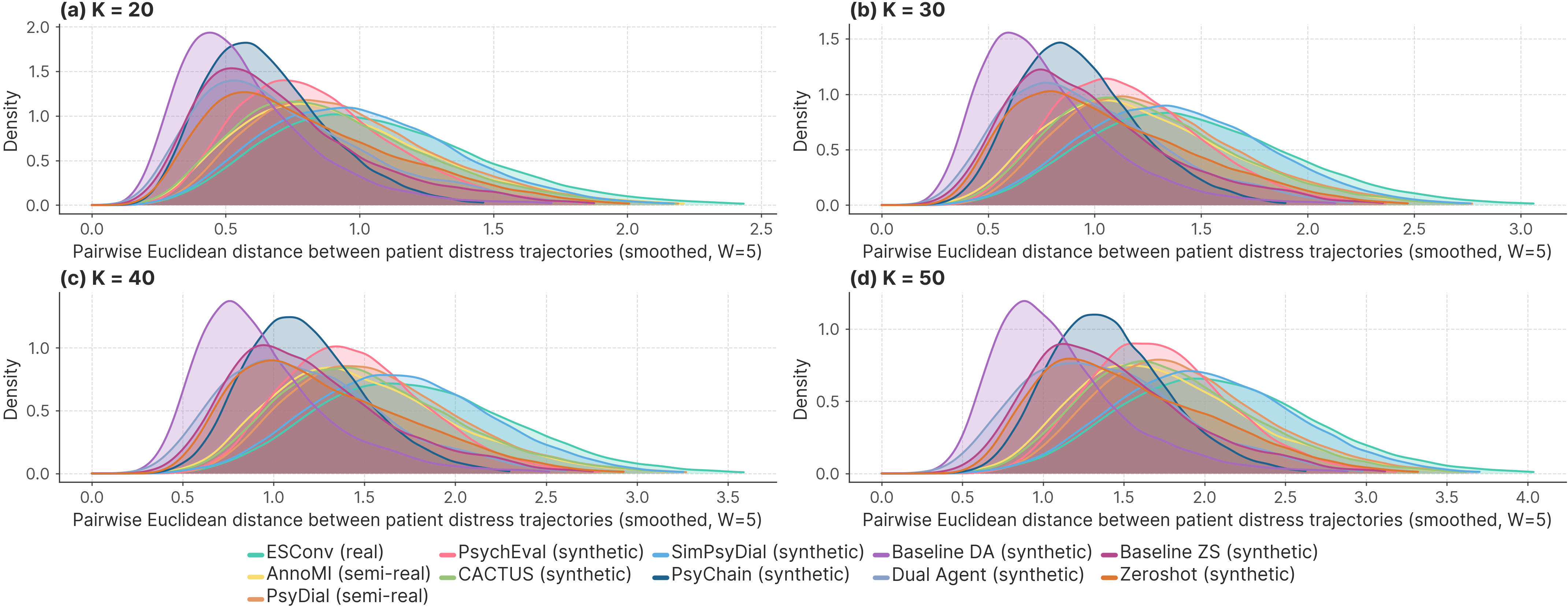}
    \caption{\textcolor{revisionblue}{
    Sensitivity of inter-conversation trajectory
    diversity to interpolation resolution.
    Moving-average smoothing windows are scaled
    proportionally with $K$ to preserve the
    original analysis procedure. Corpus rankings
    remain highly consistent across resolutions.}}
    \label{fig:grid_diversity}
\end{figure}

\paragraph{Trajectory archetype distributions.}
To isolate the effect of interpolation resolution
from variability in archetype discovery, we hold
the five archetype definitions identified at
$K=20$ fixed and reassign conversations
at each resolution.
Figure~\ref{fig:grid_archetype} shows that
the resulting corpus-level archetype
distributions remain largely stable.

For example, the proportion of conversations
classified as unresolved high distress remains
$93\%$ for Baseline DA and ranges from
$75\%$ to $78\%$ for Dual Agent.
ESConv's early-resolution proportion remains
approximately $30\%$ across all settings.
Furthermore, the corpus rankings based on
dominant-archetype share and Jensen--Shannon
divergence of archetype distributions from
ESConv remain identical across resolutions
(Kendall's $\tau=1.00$ for both measures).

These results support the stability of
archetype-based corpus comparisons under
fixed definitions. This analysis evaluates
assignment sensitivity rather than the
stability of independently refitted
archetype models.

\begin{figure}[H]
    \centering
    \captionsetup{
    labelfont={color=revisionblue},
    textfont={color=revisionblue}
}
    \includegraphics[
        width=0.75\linewidth,
        height=0.65\textheight,
        keepaspectratio
    ]{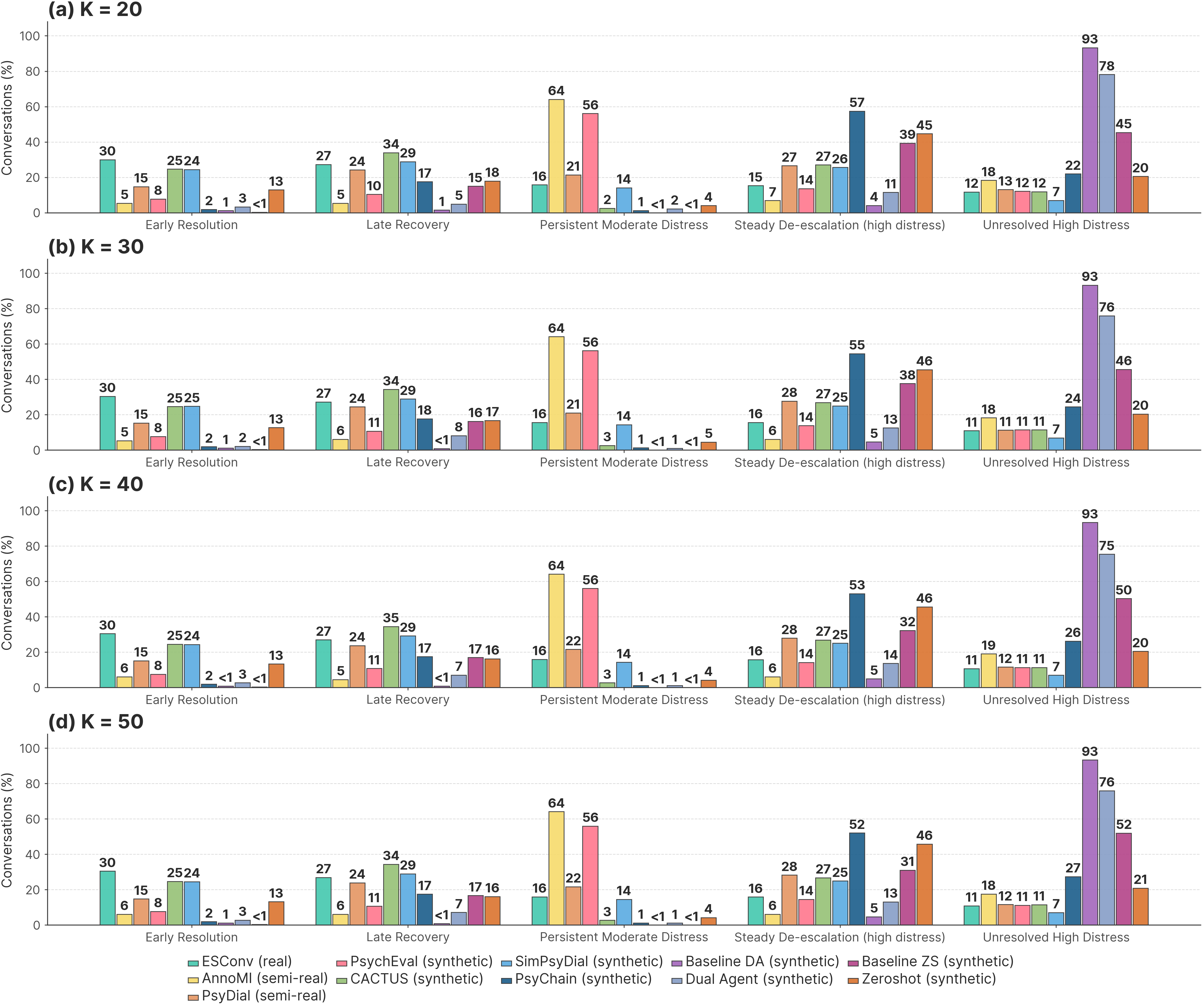}
    \caption{\textcolor{revisionblue}{
    Sensitivity of trajectory archetype
    distributions to interpolation resolution
    ($K=20,30,40,50$).
    The five archetype definitions from the
    original $K=20$ analysis are held fixed
    across resolutions.}}
    \label{fig:grid_archetype}
\end{figure}

\paragraph{Quantitative robustness summary.}
Table~\ref{tab:interpolation_robustness}
summarizes the sensitivity results across interpolation
resolutions, using $K=20$ as the reference.
Trajectory correlations and pointwise differences
quantify shape consistency, while Kendall's $\tau$
measures agreement in corpus rankings.

\begin{table}[H]
    \centering
    \captionsetup{
        labelfont={color=revisionblue},
        textfont={color=revisionblue}
    }
    \caption{\textcolor{revisionblue}{
    Quantitative sensitivity of trajectory-based
    analyses to interpolation resolution.
    All comparisons use $K=20$ as the reference.
    Minimum correlations and maximum pointwise
    differences are taken across corpora;
    Kendall's $\tau$ measures corpus-ranking
    agreement.}}
    \label{tab:interpolation_robustness}

    \small
    \setlength{\tabcolsep}{5pt}
    \renewcommand{\arraystretch}{1.08}

    {\color{revisionblue}
    \begin{tabular}{@{}lcccc@{}}
        \toprule
        \textbf{Metric}
        & $\boldsymbol{K=20}$ \textbf{(Ref.)}
        & $\boldsymbol{K=30}$
        & $\boldsymbol{K=40}$
        & $\boldsymbol{K=50}$ \\
        \midrule

        \multicolumn{5}{@{}l}{
        \textbf{\textit{Recovery trajectories}}} \\

        Minimum trajectory correlation ($r$)
            & 1.000 & 0.966 & 0.967 & 0.970 \\
        Maximum pointwise difference
            & 0.000 & 0.025 & 0.023 & 0.019 \\
        Early-to-late change ($\tau$)
            & 1.00 & 1.00 & 1.00 & 1.00 \\

        \midrule
        \multicolumn{5}{@{}l}{
        \textbf{\textit{Inter-conversation diversity}}} \\

        Fig.~3(d) diversity ranking ($\tau$)
            & 1.00 & 1.00 & 0.96 & 0.96 \\

        \midrule
        \multicolumn{5}{@{}l}{
        \textbf{\textit{Trajectory archetypes}}} \\

        Dominant-archetype share ($\tau$)
            & 1.00 & 0.96 & 0.96 & 0.96 \\
        Archetype JSD vs.\ ESConv ($\tau$)
            & 1.00 & 1.00 & 1.00 & 1.00 \\

        \bottomrule
    \end{tabular}
    }
\end{table}

\paragraph{Summary.}
Across interpolation resolutions of $K=20$--$50$,
the mean distress trajectories exhibit high shape
agreement, and recovery-trend rankings remain unchanged.
Figure~3(d) diversity rankings and dominant-archetype
share rankings remain highly consistent
(Kendall's $\tau \geq 0.96$), while archetype JSD
rankings are unchanged.
Together, these results indicate that our main
trajectory-based findings are robust to the choice
of interpolation resolution, with only minor changes
in corpus rankings and numerical values.

\endgroup
\end{document}